\documentclass[10pt,journal,compsoc]{IEEEtran}
\ifCLASSOPTIONcompsoc
  \usepackage[nocompress]{cite}
\else
  \usepackage{cite}
\fi

\usepackage{hyperref}
\usepackage{url}

\usepackage{amsmath}
\usepackage{amssymb}
\usepackage{amsthm}
\usepackage[caption=false]{subfig}
\usepackage{wrapfig}
\usepackage{graphicx}
\usepackage[noend]{algpseudocode}
\usepackage{multirow}

\newcommand{\loss}{\ell}

\newcommand{\weight}{\mathbf{w}}
\newcommand{\Weight}{\mathbf{W}}
\newcommand{\grad}{\mathbf{g}}
\newcommand{\Grad}{\mathbf{G}}

\newcommand{\ma}{\mathbf{a}}
\newcommand{\mb}{\mathbf{b}}
\newcommand{\mv}{\mathbf{v}}

\begin{document}

\title{Eva: A General Vectorized Approximation Framework for Second-order Optimization}

\author{
\IEEEauthorblockN{Lin Zhang, Shaohuai Shi,~\IEEEmembership{Member,~IEEE}, Bo Li,~\IEEEmembership{Fellow,~IEEE}\\}
\IEEEcompsocitemizethanks{\IEEEcompsocthanksitem Lin Zhang, and Bo Li are with the Department of Computer Science and Engineering, The Hong Kong University of Science and Technology, Hong Kong (email: lzhangbv@connect.ust.hk, bli@cse.ust.hk). 

\IEEEcompsocthanksitem Shaohuai Shi (corresponding author) is with the School of Computer Science and Technology, Harbin Institute of Technology, Shenzhen (email: shaohuais@hit.edu.cn).}
}



\IEEEtitleabstractindextext{%
\begin{abstract}
Second-order optimization algorithms exhibit excellent convergence properties for training deep learning models, but often incur significant computation and memory overheads. This can result in lower training efficiency than the first-order counterparts such as stochastic gradient descent (SGD). In this work, we present a memory- and time-efficient second-order algorithm named Eva with two novel techniques: 1) we construct the second-order information with the Kronecker factorization of small stochastic vectors over a mini-batch of training data to reduce memory consumption, and 2) we derive an efficient update formula without explicitly computing the inverse of matrices using the Sherman-Morrison formula. We further extend Eva to a general vectorized approximation framework to improve the compute and memory efficiency of two existing second-order algorithms (FOOF and Shampoo) without affecting their convergence performance. Extensive experimental results on different models and datasets show that Eva reduces the end-to-end training time up to $2.05\times$ and $2.42\times$ compared to first-order SGD and second-order algorithms (K-FAC and Shampoo), respectively. 
\end{abstract}

}

\maketitle

\IEEEdisplaynontitleabstractindextext

%
\IEEEpeerreviewmaketitle

\IEEEraisesectionheading{\section{Introduction}\label{sec:introduction}}

\IEEEPARstart{W}{hile} first-order optimizers such as stochastic gradient descent (SGD)~\cite{bottou1998online} and Adam~\cite{kingma2014adam} have been widely used in training deep learning models~\cite{Krizhevsky2012Alexnet,he2016deep,devlin2019bert}, these methods require a large number of iterations to converge by exploiting only the first-order gradient to update the model parameter~\cite{bottou2018optimization}. To overcome such inefficiency, second-order optimizers have been considered with the potential to accelerate the training process with a much fewer number of iterations to converge~\cite{osawa2019large,osawa2020scalable,pauloski2020convolutional,pauloski2021kaisa}. For example, our experimental results illustrate that second-order optimizers, e.g., K-FAC~\cite{martens2015optimizing}, require $\sim$50\% fewer iterations to reach the target top-1 validation accuracy of 93.5\% than SGD, in training a ResNet-110~\cite{he2016deep} model on the Cifar-10 dataset~\cite{krizhevsky2009learning} (more results are shown in Table~\ref{table:test-acc}). 

The fast convergence property of second-order algorithms benefits from preconditioning the gradient with the inverse of a matrix $C$ of curvature information. Different second-order optimizers construct $C$ by approximating different second-order information, e.g., Hessian, Gauss-Newton, and Fisher information~\cite{amari1998natural}, to help improve the convergence rate~\cite{Dennis1983NumericalMF}. However, classical second-order optimizers incur significant computation and memory overheads in training deep neural networks (DNNs), which typically have a large number of model parameters, as they require a quadratic memory complexity to store $C$, and a cubic time complexity to invert $C$, w.r.t. the number of model parameters. For example, a ResNet-50~\cite{he2016deep} model with 25.6M parameters has to store more than 650T elements in $C$ using full Hessian, which is not affordable on current devices, e.g., an Nvidia A100 GPU has 80GB memory.

To make second-order optimizers practical in deep learning, approximation techniques have been proposed to estimate $C$ with \textit{smaller matrices}. For example, the K-FAC algorithm~\cite{martens2015optimizing} uses the Kronecker factorization of two smaller matrices to approximate the Fisher information matrix (FIM) in each DNN layer, thus, K-FAC only needs to store and invert these small matrices, namely Kronecker factors (KFs), to reduce the computing and memory overheads. However, even by doing so, the additional costs of each second-order update are still significant, which makes it slower than first-order SGD. In our experiment, the iteration time of K-FAC is $2.5\times$ than that of SGD in training ResNet-50 (see Table~\ref{table:efficiency}), and the memory consumption of storing KFs and their inverse results is $12\times$ larger than that of storing the gradient. Despite the reduced number of iterations, existing second-order algorithms, such as K-FAC~\cite{martens2015optimizing}, FOOF~\cite{Benzing2022Foof}, and Shampoo~\cite{Gupta2018ShampooPS}, are \textit{not} time-and-memory efficient, as shown in Table~\ref{table:complexity}. One limitation in K-FAC and Shampoo is that they typically require dedicated system optimizations and second-order update interval tuning to outperform the first-order counterpart~\cite{osawa2019large,pauloski2020convolutional, anil2021scalable}. 

\begin{table}[!t]
    \centering
     \caption{Time and memory complexity comparison of different second-order algorithms. $d$ is the dimension of a hidden layer, $L$ is number of layers.} 
    \label{table:complexity}
    \centering
    \addtolength{\tabcolsep}{-1.2pt}
    \begin{tabular}{cccccc}
    \hline
     & Newton & K-FAC & FOOF & Shampoo & Eva \\ \hline
    Time & $O(d^6L^3)$ & $O(2d^3L)$ & $O(d^3L)$ & $O(2d^3L)$ & $O(d^2L)$ \\
    Mem. & $O(d^4L^2)$ & $O(2d^2L)$ & $O(d^2L)$ & $O(2d^2L)$ & $O(2dL)$ \\ \hline
    \end{tabular}
\end{table}
 
To address the above limitations, we propose a novel second-order training algorithm, called Eva, which introduces a matrix-free approximation to the second-order matrix to precondition the gradient. Eva not only requires much less memory to estimate the second-order information, but it also does not need to explicitly compute the inverse of the second-order matrix, thus eliminating the intensive computations required in existing methods. Specifically, we propose two novel techniques in Eva. First, for each DNN layer, we exploit the Kronecker factorization of two small stochastic vectors, called Kronecker vectors (KVs), over a mini-batch of training data to construct a rank-one matrix to be the second-order matrix $C$ for preconditioning. KVs require only a sublinear memory complexity w.r.t. the model size, which is much smaller than the linear memory complexity in existing second-order algorithms like storing KFs in K-FAC~\cite{martens2015optimizing}, FOOF~\cite{Benzing2022Foof}, and Shampoo~\cite{Gupta2018ShampooPS} (see Table~\ref{table:complexity}), or gradient copies in M-FAC~\cite{mfac2021frantar}. Second, we derive a new update formula to precondition the gradient by implicitly computing the inverse of the constructed Kronecker factorization using the Sherman–Morrison formula~\cite{sherman1950adjustment}. The new update formula takes only a linear time complexity; it means that Eva is much more time-efficient than existing second-order optimizers which normally take a superlinear time complexity in inverting matrices (see Table~\ref{table:complexity}). In addition, we generalize our Eva algorithm to a vectorized approximation framework so that we can treat Eva as a vectorized version of K-FAC. From this perspective, our vectorized approximation can be used to optimize the high compute and memory overheads of more existing second-order methods, including FOOF and Shampoo. Specifically, we derive two novel vectorized second-order algorithms (Eva-f and Eva-s), and they enjoy the same system advantages like Eva, without losing convergence performance. 

We conduct extensive experiments to illustrate the effectiveness and efficiency of Eva compared to widely used first-order (SGD, Adagrad, and Adam) and second-order (K-FAC, Shampoo, and M-FAC) optimizers on multiple deep models and datasets. The experimental results show that 1) Eva outperforms first-order optimizers -- achieving higher accuracy under the same number of iterations or reaching the same accuracy with fewer number of iterations, and 2) Eva generalizes very closely to other second-order algorithms such as K-FAC while having much less iteration time and memory footprint. Specifically, in terms of per-iteration time performance, Eva only requires an average of $1.14\times$ wall-clock time over first-order SGD, while K-FAC requires $3.47\times$ in each second-order update. In term of memory consumption, Eva requires almost the same memory as first-order SGD, which is up to $31\%$ and $45\%$ smaller than second-order Shampoo and K-FAC respectively. In term of the end-to-end training performance, experimental results on different training benchmarks show that Eva reduces the training time up to $2.05\times$, $1.58\times$, and $2.42\times$ compared to SGD, K-FAC, and Shampoo respectively. Furthermore, our experiemntal results show that the idea of Eva can be generalized to two more second-order methods (FOOF and Shampoo) to improve their system efficiency. 

In summary, our contributions are as follows: 
(1) We propose a novel efficient second-order optimizer Eva via Kronecker-vectorized approximation. Eva has a sublinear memory complexity, and it reduces each second-order update cost to linear time complexity. 
(2) We implement Eva atop PyTorch framework and provide easy-to-use APIs so that users can adopt Eva by adding several lines of code in their training scripts.
(3) We conduct extensive experiments to validate that Eva can converge faster than SGD, and it is more system efficient than K-FAC and Shampoo. Therefore, Eva is capable of improving end-to-end training performance. 

As an extension to our prior work~\cite{lin2022eva}, we summarize additional contributions as follows:     
(1) We introduce three commonly used second-order optimization methods (K-FAC, FOOF, and Shampoo) in a unified Kronecker factorization based approximation manner. (2) We generalize the idea of Eva and propose a vectorized approximation framework. It is applied to FOOF and Shampoo to improve their compute and memory efficiency for training DNNs. (3) We provide two explanations to understand our vectorized approximation, and conduct experiments to validate the effectiveness and efficiency of our vectorized algorithms Eva-f and Eva-s. 

\section{Background} \label{sec:background}

\begin{figure}[!t]
    \centering
    \includegraphics[width=0.99\linewidth]{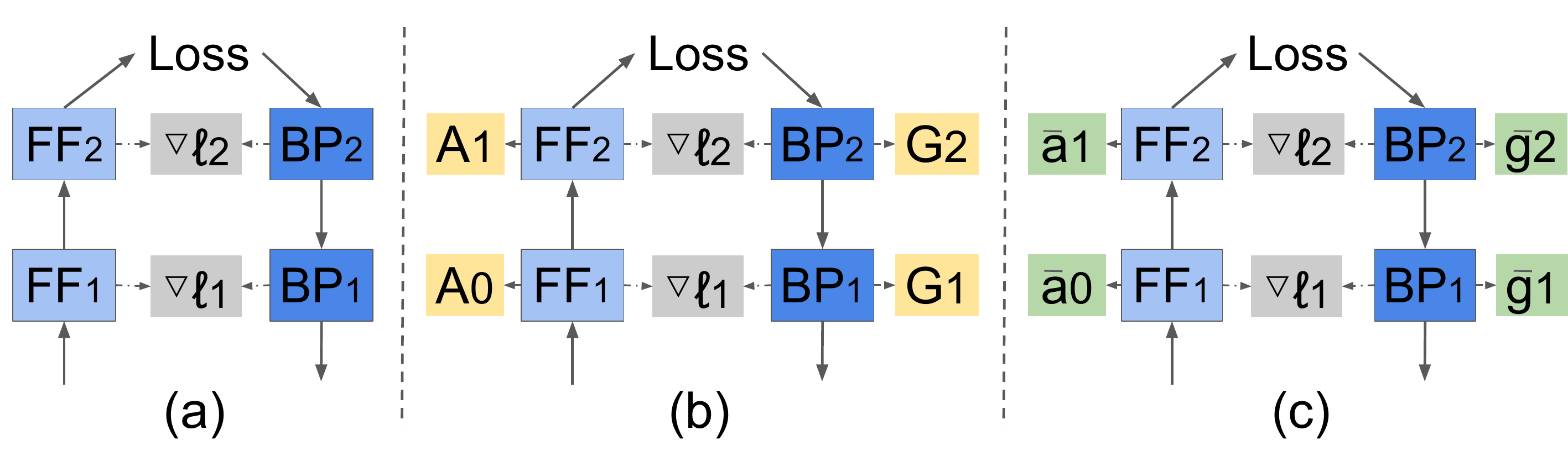}
    \caption{Examples of different optimization algorithms in a two-layer DNN model: (a) SGD, (b) K-FAC, (c) Eva. } 
    \label{fig:kfac-examples}
\end{figure}

In this section, we give preliminaries on useful notations and definitions, as well as the Sherman–Morrison formula. Then we present the background of first-order SGD and other relevant second-order optimization algorithms. 

\subsection{Preliminaries}
We start from the supervised learning, in which a DNN model is trained by randomly going through the dataset many times (i.e., epochs) to minimize a loss function. The loss function measures the average distance between model predictions and ground-truth labels. Given a DNN with $L$ learnable layers, we use $\weight$ to denote the total model parameter to be trained, which typically consists of a set of parameter matrices $\{W_{l}\}_{l=1}^L$, or a set of parameter tensors $\{\Weight_{l}\}_{l=1}^L$. The bias parameters are either concatenated into weights, or skipped during the preconditioning. 

\begin{table}[!ht]
    \centering
    \caption{Notations.}
    \label{table:notation}
    \begin{tabular}{c|l}
    \hline
    Name & Description \\ \hline \hline
    $\weight$, W, $\Weight$ & model parameter vector, matrix, and tensor \\ \hline
    $\grad$, G, $\Grad$ & gradient vector, matrix, and tensor w.r.t. parameter \\ \hline
    $C$ & curvature information matrix \\ \hline
    $A$, $B$ & input activation, and pre-activation output gradient \\ \hline
    $Q$, $R$, $M$ & different Kronecker factors (KFs) \\ \hline
    $\bar{\ma}$, $\bar{\mb}$, $\mv$ & different Kronecker vectors (KVs) \\ \hline
    $\alpha$ & learning rate \\ \hline
    $\gamma$ & damping value \\ \hline
    \end{tabular}
\end{table}

\textbf{Notations and definitions.} We summarize some frequently used notations in Table~\ref{table:notation}. We use lowercase and bold lowercase letters to denote scalars and vectors, and uppercase and bold uppercase letters to denote matrices and tensors, respectively. 

For a model gradient matrix $G$, it can be flattened into a vector $\grad$ by concatenating rows. Note it differs from the notation of concatenating columns used in our previous work~\cite{lin2022eva}, which results in a slightly different property on Kronecker product of $\otimes$. That is, given two symmetric matrices $A$ and $B$, $(A \otimes B)\grad$ is equivalent to $AGB$. The notation $A \succeq B$ means that $A-B$ is a positive semi-definite matrix. If $A \succeq A'$ and $B \succeq B'$, then $A \otimes B \succeq A' \otimes B'$. 

For an order-k gradient tensor $\Grad$ of dimensions $d_1 \times \cdots \times d_k$, it can be reshaped into a matrix via $\text{mat}_i(\Grad)$ with the shape of $(d_i, d_{-i})$, where $d_i$ is the $i$-th dimension, and $d_{-i}=\prod_{j\neq i}d_j$; or it can be flattened into a vector $\grad$. The product of a tensor $\Grad$ with a symmetric matrix $M_i$ is defined by $\text{mat}_i(\Grad \times_i M_i) = M_i \text{mat}_i(\Grad)$, and $\Grad \times_i M_i$ is a tensor with the same shape of $\Grad$. This gives a useful property on Kronecker product, that is, $(M_1 \otimes \cdots \otimes M_k) \grad$ is equivalent to $\Grad \times_1 M_1 \times_2 \cdots \times_k M_k$, given $k$ symmetric matrices. 

\textbf{Sherman–Morrison formula.} Given an invertible square matrix $A$, the Sherman–Morrison formula~\cite{sherman1950adjustment} is 
\begin{equation}\label{eq:sm-invert}
    (A + \mathbf{u}_1 \mathbf{u}^T_2)^{-1} = A^{-1} - \frac{A^{-1} \mathbf{u}_1\mathbf{u}^T_2 A^{-1}}{1 + \mathbf{u}^T_2 A^{-1}\mathbf{u}_1}, 
\end{equation}
where $\mathbf{u}_1$ and $\mathbf{u}_2$ are two vectors. If the inverse of $A$ is already known, it provides a cheap way to compute the inverse of $A$ perturbed by a rank-one update $\mathbf{u}_1 \mathbf{u}_2^T$. In this work, we use Sherman–Morrison formula to compute the inverse of the curvature matrix with damping efficiently in Eva algorithms. 
\subsection{First-order and second-order algorithms}
\textbf{SGD.} The stochastic gradient descent (SGD) algorithm and its variants (e.g., Adam) with first-order gradient information are the main optimizers for training DNN models. In each iteration with a mini-batch of data, SGD updates the model parameter as follows: 
\begin{equation}\label{eq:sgd}
    \weight_l^{(t+1)} \leftarrow \weight_l^{(t)} + \Delta \weight_l \quad\text{and}\quad \Delta \weight_l = - \alpha \grad_l,
\end{equation}
where $\weight_l, \grad_l$ are model parameter and first-order gradient at layer $l$, respectively. $\alpha > 0$ is the learning rate. SGD typically takes a large number of iterations to converge as it only utilizes the first-order gradient to optimize the model~\cite{bottou2018optimization}. 

\textbf{Preconditioned SGD.} Compared to SGD, second-order algorithms precondition the gradient by the inverse of the curvature matrix, i.e., $C^{-1}\grad$, for model update. The straight-forward way is to use the Hessian to precondition the gradient using the Newton method, but it takes extremely high time and memory complexity as shown in Table~\ref{table:complexity}. Due to the deep structure of DNNs, storing and inverting full model's $C$ are expensive, which motivates block-diagonalization to alleviate the memory and computation complexity. That is, we precondition the gradient layer-wisely to update the model parameter as follows:  
\begin{equation}\label{eq:psgd}
    \Delta \weight_l = - \alpha C_l^{-1} \grad_l, 
\end{equation}
where the full model's curvature is represented by a block-diagonal matrix $C = \text{diag}(C_1, \cdots, C_L)$. Based on it, many approximation approaches have been proposed to construct $C_l$ with smaller matrices, such as K-FAC~\cite{martens2015optimizing}, FOOF~\cite{Benzing2022Foof}, and Shampoo~\cite{Gupta2018ShampooPS}. They have been successfully applied in large-scale training to achieve comparable performance over SGD~\cite{osawa2019large, osawa2020scalable,pauloski2020convolutional,pauloski2021kaisa,anil2021scalable}. 

In this work, we introduce K-FAC, FOOF, and Shampoo from a unified Kronecker factorization based approximation perspective as follows. 


\textbf{K-FAC.} The K-FAC algorithm~\cite{martens2015optimizing} adopts the Fisher information matrix (FIM) as $C_l$ for layer $l$ in a DNN and approximates $C_l$ with the Kronecker product of two smaller matrices~\cite{martens2015optimizing,grosse2016kronecker}~\footnote{In this paper, we focus on the oft-used K-FAC with \textit{empirical} FIM. }. For example, for a linear layer $A_l = \phi(W_l A_{l-1})$, in which $\phi$ is an element-wise non-linear activation function, the FIM can be approximated via $C_{l}=Q_{l} \otimes R_{l}$, where
\begin{equation}\label{eq:kfac-approx}
    Q_{l} = B_l B_l^T \quad \text{ and }\quad R_{l} = A_{l-1}A_{l-1}^T.
\end{equation}
$Q_{l} \in \mathbb{R}^{d_{l} \times d_{l}}$ and $R_{l} \in \mathbb{R}^{d_{l-1} \times d_{l-1}}$ are symmetric matrices, called Kronecker factors (KFs). $A_{l-1} \in \mathbb{R}^{d_{l-1} \times n}$ is the batched input activation of layer $l$ (i.e., output of layer $l-1$) and $B_l \in \mathbb{R}^{d_{l} \times n}$ is the batched pre-activation output gradient of layer $l$, and $n$ is the number of samples. $\otimes$ denotes the Kronecker product. KFs are used to precondition the gradient with a damping parameter $\gamma >0$ via $(Q_{l}\otimes R_{l}+\gamma I)^{-1}\grad_l$. Formally, we have 
\begin{equation}\label{eq:kfac}
    \Delta W_l = -\alpha (Q_{l} + \gamma_L I)^{-1} G_l(R_{l} + \gamma_R I)^{-1},
\end{equation}
where $W_l, G_l \in \mathbb{R}^{d_l \times d_{l-1}}$ are model parameter and first-order gradient matrices at layer $l$, respectively. $\gamma_L = \sqrt{\gamma}/\pi_{l}$ and $\gamma_R=\pi_{l} \sqrt{\gamma}$ are damping terms, and the detailed derivation of K-FAC is given in the Appendix of~\cite{lin2022eva}. An example of a 2-layer DNN is shown in Fig.~\ref{fig:kfac-examples}(a) and Fig.~\ref{fig:kfac-examples}(b) to demonstrate the difference between SGD and K-FAC. Compared to the training process of SGD, K-FAC extra constructs two KFs, which have a memory complexity of $O(2d^2)$, to approximate FIM and computes its inverse, which has a time complexity of $O(2d^3)$, to precondition the gradient in each layer. 

\textbf{FOOF.} The FOOF~\cite{Benzing2022Foof} is a gradient descent gradient algorithm on neurons, and it can be viewed as applying the right-sided preconditioning of K-FAC as follows: 
\begin{equation}\label{eq:foof}
    \Delta W_l = -\alpha G_l(R_{l} + \gamma I)^{-1},
\end{equation}
where $R_l = A_{l-1}A_{l-1}^T$ is the same KF, derived from batched input of layer $l$, and $\gamma > 0$ is the damping value. Compared to K-FAC, FOOF adopts the preconditioned matrix as $C_l = I \otimes R_l$, which only requires constructing and inverting the KF of $R_l$. Thus, it can reduce the memory complexity to $O(d^2)$, and reduce the time complexity of computing its inverse to $O(d^3)$ in each layer.  

\textbf{Shampoo.} The Shampoo~\cite{Gupta2018ShampooPS} is a full-matrix adaptive algorithm that preconditions the gradient tensor with gradient statistics matrices. Specifically, gradient statistics are constructed by
\begin{equation}
    M_{l,i} = \text{mat}_i(\Grad_l) \text{mat}_i(\Grad_l)^T,
\end{equation}
where $\Grad_l$ is the gradient tensor of layer $l$. For each dimension of the gradient tensor, one can reshape $\Grad_l$ into a matrix with the size of $(d_i, d_{-i})$ via $\text{mat}_i(\Grad_l)$, where $d_i$ is the $i$-th dimension of $\Grad_l$. Based on it, one can calculate the gradient statistic $M_{l,i}$ of layer $l$ and dimension $i$. The gradient statistic can be regarded as the KF, since the preconditioner matrix used in Shampoo is given by $C_l = \otimes_{i=1}^k M_{l,i}$, where $k$ is the number of dimensions of the gradient tensor $\Grad_l$. 

The KFs in each layer are used to precondition the gradient via $C_{l}^{-1/2k}\grad_l$, which gives the following update formula: 
\begin{equation}\label{eq:shampoo}
    \Delta \Weight_l = -\alpha \Grad_l \times_1 (M_{l,1}+ \gamma I)^{-\frac{1}{2k}} \times_2 \cdots \times_k (M_{l,k}+ \gamma I)^{-\frac{1}{2k}}, 
\end{equation}
where $M_{l, i}$ ($i=1, \cdots, k$) are KFs, $k$ is the number of dimensions of the gradient tensor $\Grad_l$. $\times_i$ is the tensor-matrix multiplication operator over $i$-th dimension, and $\gamma$ is the damping value. The update formula of Shampoo is very similar to K-FAC, except that it preconditions the gradient tensor rather than gradient matrix, and it computes the $2k$-th root inverse of KFs rather than inverse of KFs. The memory complexity and time complexity to construct and invert KFs in Shampoo are $O(k d^2)$, and $O(k d^3)$, respectively. For a fair comparison, we assume gradient tensors are matrices like in K-FAC and FOOF, and it gives the memory complexity of $O(2 d^2)$, and the time complexity of $O(2d^3)$. 

In summary, the time and memory complexity of K-FAC, FOOF, and Shampoo is listed in Table~\ref{table:complexity}. Motivated by the fact that existing second-order methods have high time and/or memory complexity in each second-order update, in this work, we present a novel second-order optimizer, Eva, with much lower time and memory complexity. 

\textbf{Additional approaches.} There exist some other second-order optimizers for DNN training. For example, Hessian-free (HF) optimization~\cite{martens2010deep} enables Newton method into large neural network training, which does not construct the explicit Hessian matrix, but requires extra feed-forward and back-propagation passes to calculate the Hessian-vector products with an iterative conjugate gradient (CG) approach. The time cost of HF optimization is dozens of times (i.e., the number of CG iterations) than SGD. Following HF optimization, Adahessian~\cite{yao2021adahessian} requires only one extra feed-forward and back-propagation pass to estimate Hessian diagonal, but the estimation can be very inaccurate. Sophia~\cite{liu2023sophia} equips Adahessian with element-wise clipping to handle heterogeneous curvatures in language model pre-training. M-FAC~\cite{mfac2021frantar} is a matrix-free algorithm that utilizes matrix-vector products with many gradient copies. M-FAC does not invert any matrix but requires $m$ copies of gradient to estimate FIM, which brings time and memory costs of $O(md^2L)$. $m$ is typically suggested to be as high as 1024~\cite{mfac2021frantar}. 

\section{Kronecker Factored Approximation with Small Vectors} \label{sec:method}

\subsection{Eva: vectorize K-FAC}
Following standard feed-forward (FF) and back-propagation (BP) processes, as shown in Fig.~\ref{fig:kfac-examples}(c), we propose to construct the second-order curvature matrix $C_l$ for layer $l$ using the Kronecker product of two small vectors $\bar{\ma}_{l-1}$ and $\bar{\mb}_{l}$, i.e.,
\begin{equation}\label{eq:fast-kfac-approx}
    C_l = (\bar{\mb}_{l} \bar{\mb}_{l}^T) \otimes (\bar{\ma}_{l-1} \bar{\ma}_{l-1}^T) = (\bar{\mb}_{l} \otimes \bar{\ma}_{l-1})  (\bar{\mb}_{l} \otimes \bar{\ma}_{l-1})^T,
\end{equation}
where
\begin{equation}\label{eq:kv}
    \bar{\ma}_{l-1} = \text{mean-col}(A_{l-1})
    \quad \text{ and }\quad \bar{\mb}_{l} = \text{mean-col}(B_l). 
\end{equation}
$\bar{\ma}_{l-1}$ and $\bar{\mb}_{l}$ are average activation and pre-activation gradient vectors at layer $l$, respectively, over a mini-batch of data. We call $\bar{\ma}_{l-1}$ and $\bar{\mb}_{l}$ as Kronecker vectors (KVs), whose dimensions are irrelevant with the batch size. Thus, we just need to store inexpensive KVs to construct $C_l$, which is much more memory-efficient than storing the whole matrix like K-FAC or Shampoo. The preconditioner of Eva becomes $(C_l+\gamma I)^{-1}$.

However, it still needs to compute the inverse of the damped $C_l$, which is computation-expensive in training DNNs. From Eq.~\ref{eq:fast-kfac-approx}, $C_l$ is a rank-one matrix. Using this good property, we can implicitly compute its inverse efficiently using the Sherman-Morrison formula~\cite{sherman1950adjustment}. Specifically, 
\begin{align}\label{eq:fast-kfac-inverse}
    (C_l + \gamma I)^{-1} &= ((\bar{\mb}_{l} \otimes \bar{\ma}_{l-1})  (\bar{\mb}_{l} \otimes \bar{\ma}_{l-1})^T + \gamma I)^{-1} \\
    &= \frac{1}{\gamma}(I - \frac{(\bar{\mb}_{l} \bar{\mb}_{l}^T) \otimes (\bar{\ma}_{l-1} \bar{\ma}_{l-1}^T)}{\gamma + (\bar{\ma}_{l-1}^T\bar{\ma}_{l-1})(\bar{\mb}_{l}^T\bar{\mb}_{l})}).
\end{align}
Based on the above preconditioner, we can derive the update formula of Eva to be
\begin{equation}\label{eq:eva}
     \Delta W_l = - \frac{\alpha}{\gamma} \Big(G_l - \frac{\bar{\mb}_{l}^T G_l \bar{\ma}_{l-1}}{\gamma + (\bar{\ma}_{l-1}^T\bar{\ma}_{l-1})(\bar{\mb}_{l}^T\bar{\mb}_{l})} \bar{\mb}_{l} \bar{\ma}_{l-1}^T \Big). 
\end{equation}
The detailed derivation of Eva is in the Appendix of~\cite{lin2022eva}. Compared to SGD, the precondition process of Eva changes the gradient direction with compensating in the direction of $\bar{\mb}_{l} \bar{\ma}_{l-1}^T$ and scales the step size by $1/\gamma$. 

\textbf{Running average of KVs.}
In Eva, KVs are calculated over a mini-batch of data for constructing the second-order preconditioner by their Kronecker product. For a large-scale training dataset, the approximation can be stabilized by a long-term run of estimation using the whole data. Thus, we use the running average strategy which is also commonly used in Adam~\cite{kingma2014adam} with 1st and 2nd moments, or K-FAC~\cite{martens2015optimizing} with KFs, that is
\begin{align}
    \bar{\ma}_{l-1}^{(t)} &\leftarrow \xi \bar{\ma}_{l-1}^{(t)} + (1-\xi) \bar{\ma}_{l-1}^{(t-1)}, \\
    \bar{\mb}_{l}^{(t)} &\leftarrow \xi \bar{\mb}_{l}^{(t)} + (1-\xi) \bar{\mb}_{l}^{(t-1)},
\end{align}
where $\xi \in (0, 1]$ is the running average parameter in iteration $t$, $\bar{\ma}_{l-1}^{(t)}$ and $\bar{\mb}_{l}^{(t)}$ in the right-hand side are new KVs calculated during each FF and BP computation and they are used to update the state of KVs. 

\textbf{Preconditioned gradient clipping.} After the gradient is preconditioned with KVs, the magnitude of preconditioned gradient is typically much larger than that of the original gradient, which could cause divergence. To prevent exploding the preconditioned gradient, we clip the preconditioned gradient when the KL size is higher than a threshold~\cite{pauloski2020convolutional}. The KL size is defined by $\textbf{p}_l^T \grad_l$, where $\textbf{p}_l$ and $\grad_l$ are preconditioned gradient and gradient, respectively, at layer $l$. Given an $L$-layer DNN model, KL clipping is to scale each preconditioned gradient by a factor of
\begin{equation}~\label{eq:kl-clip}
    \nu_{KL} = \min \Big(1, \sqrt{\frac{\kappa}{\alpha^2 \sum_{l=1}^L \textbf{p}_l^T \grad_l}} \Big),
\end{equation}
where $\kappa > 0$ is the threshold for KL clipping, and $\alpha$ is the learning rate. As KL clipping is used in the practise of K-FAC~\cite{pauloski2020convolutional}, we also apply it into Eva for a fair comparison. However, KL clipping is coupled with learning rate schedule, and it requires an extra threshold hyper-parameter. We will present hyper-parameter free methods later. 

\textbf{Complexity analysis.} The extra time costs of Eva come from constructing KVs and preconditioning the gradients with KVs, apart from sharing the same FF, BP, and update computations as SGD. As the overhead of estimating KVs can be ignored, the main time cost of Eva is multiple vector multiplications for preconditioning, which can be denoted as $O(d^2L)$. It means Eva has a time complexity that is \textit{linear} to the total number of parameters, which is much smaller than the superlinear complexity in K-FAC and Shampoo as shown in Table~\ref{table:complexity}. Besides, the memory complexity of Eva is $O(2dL)$ for storing KVs, which is \textit{sublinear} to the total number of parameters. In summary, Eva has very little extra time and memory costs in each second-order update compared to first-order SGD, but it enjoys the fast convergence property of second-order K-FAC. 

\subsection{Theoretical understanding}\label{subsec:theory}

\begin{figure}
    \centering
    \includegraphics[width=0.99\linewidth]{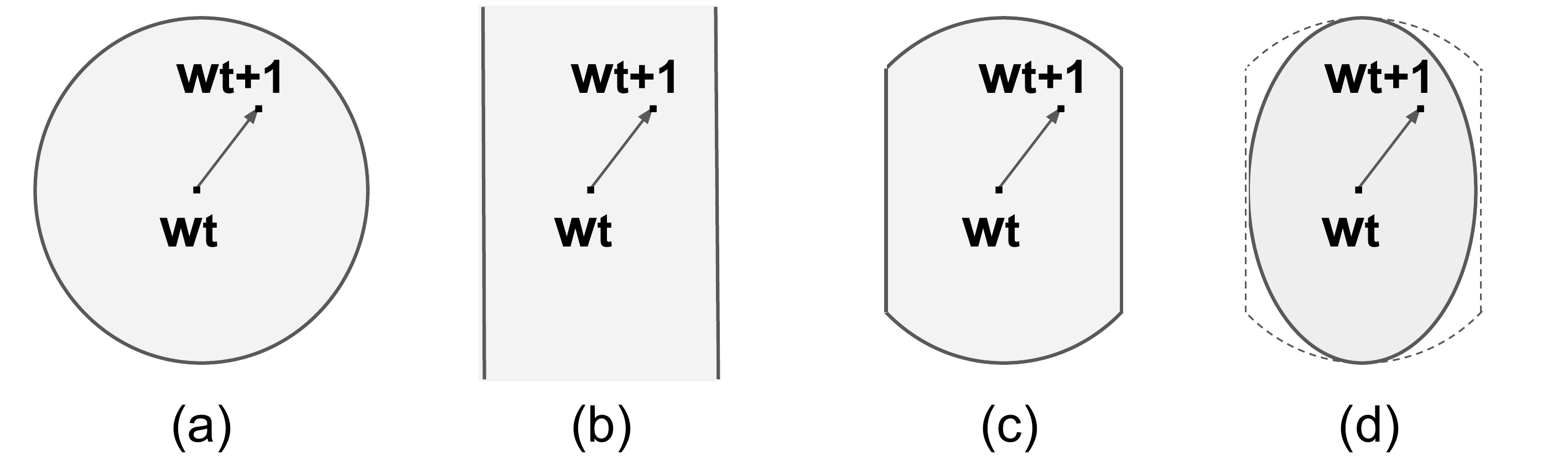}
    \caption{Trust regions of different proximal constraints: (a) ball region, (b) strip region, (c) trust region of Eva, (d) trust region of K-FAC. }
    \label{fig:trust-region}
\end{figure}

\textbf{Trust-region optimization.}
The trust-region optimization algorithm~\cite{Yuan2015RecentTR} can be formulated as 
\begin{align}
    & \weight^{(t+1)} = \min_{\weight} \loss(\weight), \\
    & \text{s.t. }  \rho(\weight, \weight^{(t)}) \le \epsilon, 
\end{align}
where $\weight^{(t)}$ and $\weight^{(t+1)}$ are current and next parameter points, respectively. $\rho(\weight, \weight^{(t)})$ is the proximal function to measure the distance between $\weight$ and $\weight^{(t)}$ (the smaller the closer). Trust-region optimization treats each update step as finding the next parameter point that minimizes the loss function and is close to the current parameter point (i.e., located in a trust region of $\{\weight: \rho(\weight, \weight^{(t)}) \le \epsilon\}$, where $\epsilon > 0$ is the threshold). Different proximal functions correspond to different trust regions. For example, the euclidean distance, $\rho(\weight, \weight^{(t)}) = ||\weight - \weight^{(t)}||^2=\Delta \weight^T \Delta \weight$, gives a ball region as shown in Fig.~\ref{fig:trust-region}(a). In particular, if the trust region is defined by $\rho(\weight, \weight^{(t)}) = \Delta\weight^T C \Delta\weight$, the optimal solution of trust-region optimization is given by $\Delta \weight=C^{-1} \grad$ (more details are in the Appendix of~\cite{lin2022eva}). It means many second-order algorithms can be understood from trust-region optimization with different trust regions.

\textbf{Trust region of Eva.} In Eva, we have $C_l = \mv_l\mv_l^T + \gamma I$, where $\mv_l=\bar{\mb}_{l} \otimes \bar{\ma}_{l-1}$. It consists of two trust regions: the first part is $\Delta \weight_l^T (\mv_l\mv_l^T)\Delta \weight_l = (\Delta \weight_l^T \mv_l)^2$, as shown in Fig.~\ref{fig:trust-region}(b), which limits the update size along the $\mv_l$ direction; and the second part is the ball region. Therefore, it indicates that the trust region of Eva is an intersection between the ball region and the strip region, as shown in Fig.~\ref{fig:trust-region}(c).

\textbf{Relation to K-FAC.} In K-FAC, we have $Q_l \otimes R_l = B_lB_l^T \otimes A_{l-1}A_{l-1}^T$. Since we have the following identity:
\begin{equation}\label{eq:A-a}
    \frac{1}{n} A A^T = \bar{\ma}\bar{\ma}^T + \frac{1}{n} \sum_{i=1}^n (\ma_i-\bar{\ma})(\ma_i-\bar{\ma})^T.  
\end{equation}
It indicates that $AA^T \succeq \bar{\ma}\bar{\ma}^T$, given any matrix $A$, and its average vector over columns $\bar{\ma}=\sum_{i=1}^n \ma_i /n$. Now consider the relation between K-FAC's KFs and Eva's KVs, we have $B_lB_l^T \otimes A_{l-1}A_{l-1}^T \succeq (\bar{\mb}_{l} \bar{\mb}_{l}^T) \otimes (\bar{\ma}_{l-1} \bar{\ma}_{l-1}^T)$. In other words, 
for a given threshold $\epsilon$, the trust region of Eva is larger than K-FAC as shown in Fig.~\ref{fig:trust-region}(d). Thus, the update step of K-FAC to minimize the loss is more conservative than Eva. 

\subsection{Implementation}
We implement Eva atop PyTorch. Following~\cite{pauloski2020convolutional,pauloski2021kaisa}, we build a preconditioner to estimate KVs and precondition the gradients before performing the standard SGD optimizer. Our preconditioner supports Linear and Conv2D layers, and parameters at other layers (e.g., BatchNorm2d) will be updated by SGD without preconditioning. To construct KVs at supported layers, we register forward pre-hooks and backward-hooks to capture the activations and pre-activation gradients during the feed-forward and back-propagation computations, respectively. These intermediate values are used to estimate new KVs and update the running average states. Then we use stored KVs to precondition the gradients layer-wisely, and perform gradient clipping on preconditioned gradients, before they are used to update the model parameters in SGD. 

To support data parallelism, we implement our Eva preconditioner to communicate KVs at each worker via \textit{all-reduce} primitives. The communication of KVs is very efficient as the data volume of KVs is sublinear to the number of gradients, and small KVs can be merged to be communicated together via the tensor fusion technique supported by the distributed training framework Horovod~\cite{sergeev2018horovod}. The aggregated KVs are then used to precondition the aggregated gradients on all workers. Unlike distributed K-FAC~\cite{osawa2019large, shi2021accelerating}, distributed Eva does not need to assign matrix-inversion tasks at different layers into different workers, and it also does not need to use stale FIM to skip the precondition of many iterations (e.g., update KFs every 50 iterations). Instead, it is memory- and time-efficient to construct KVs and precondition the gradients on all workers during the whole training process. 

\section{A Vectorized Approximation Framework}
In this section, we generalize the idea of Eva as a vectorized approximation framework, and apply it into two more second-order optimization algorithms FOOF and Shampoo. 

Similar to K-FAC, FOOF and Shampoo use KFs to approximate the curvature information via Kronecker factorization, as shown in Table~\ref{tab:evas}. Each KF is a symmetric matrix, and it is typically constructed as $M=XX^T$. In our vectorized approximation framework, each $X$ (e.g., $A_{l-1}$, $B_l$ and $\text{mat}_i(\Grad_l)$) is vectorized by simply taking the average of columns (i.e., $\mv=\text{mean-col}(X)$). Instead, we use these Kronecker vectors (KVs) to construct a novel curvature matrix $C_l$, and precondition the gradient layer-wisely via $(C_l+\gamma I)^{-1} \grad_l$, where the damped curvature can be converted efficiently via Sherman-Morrison formula without any explicit inversion operation. 

\begin{table}[!ht]
    \centering
    \caption{Curvature information approximations of different algorithms. }
    \label{tab:evas}
    \begin{tabular}{c|l}
    \hline
    Algorithm & Curvature information approximation \\\hline
    K-FAC & $C_l=B_lB_l^T \otimes A_{l-1}A_{l-1}^T$ \\
    Eva & $C_l=(\bar{\mb}_{l} \bar{\mb}_{l}^T) \otimes (\bar{\ma}_{l-1} \bar{\ma}_{l-1}^T)$ \\\hline
    FOOF & $C_l=I \otimes A_{l-1}A_{l-1}^T$ \\
    Eva-f & $C_l=I \otimes (\bar{\ma}_{l-1} \bar{\ma}_{l-1}^T)$ \\\hline
    Shampoo & $C_l=\otimes_{i=1}^k \text{mat}_i(\Grad_l)\text{mat}_i(\Grad_l)^T$ \\
    Eva-s & $C_l=\otimes_{i=1}^k \mv_{l,i}\mv_{l,i}^T$ \\\hline
    \end{tabular}
\end{table}

\subsection{Eva-f: vectorize FOOF}
To vectorize FOOF, we only need to replace the KF of $R_l=A_{l-1}A_{l-1}^T$ by $\bar{\ma}_{l-1}\bar{\ma}_{l-1}^T$. The corresponding update formula is given as follows: 
\begin{align}\label{eq:eva-f}
    \Delta W_l &= -\alpha G_l(\bar{\ma}_{l-1}\bar{\ma}_{l-1}^T+\gamma I)^{-1}, \\ 
    &= - \frac{\alpha}{\gamma}\Big(G_l - \frac{G_l\bar{\ma}_{l-1}\bar{\ma}_{l-1}^T}{\gamma + \bar{\ma}_{l-1}^T\bar{\ma}_{l-1}} \Big). 
\end{align}

\textbf{Preconditioned gradient normalization.} To avoid the efforts of hyper-parameter tuning with KL clipping, we propose to use KL normalization for Eva-f. This is based on the observation that KL clipping is equal to normalize the preconditioned gradient by its KL size when the threshold is small enough. Specifically, the preconditioned gradients are KL-normalized, i.e., $\textbf{p}_l/\sqrt{\sum_{l=1}^L \textbf{p}_l^T\grad_l}$, before they are used to update model parameters. Thus, it requires no extra hyper-parameter except learning rate to control the update step size, which is as convenient as SGD. 

\subsection{Eva-s: vectorize Shampoo}
In Shampoo, we vectorize the curvature matrix i.e.
\begin{equation}
    C_l = \otimes_{i=1}^k (\mv_{l,i}\mv_{l,i}^T) = (\otimes_{i=1}^k \mv_{l,i})(\otimes_{i=1}^k \mv_{l,i})^T, 
\end{equation}
where $\mv_{l,i}=\text{mean}_{-i}(\Grad_l)$, and $\text{mean}_{-i}$ takes average on the gradient tensor over all dimensions except $i$. 
Since $C_l$ is a rank-one matrix, we can invert $C_l + \gamma I$ very efficiently, and the formula of $(C_l + \gamma I)^{-1}\grad_l$ can be re-written as precondition the gradient tensor as follows:
\begin{equation}
    \Delta \Weight_l = - \frac{\alpha}{\gamma}\Big(\Grad_l - \frac{\Grad_l \times_1 \mv_{l,1}\mv_{l,1}^T \times_2 \cdots \times_k \mv_{l,k}\mv_{l,k}^T }{\gamma + \prod_{i=1}^k \mv_{l,i}^T \mv_{l,i}} \Big). 
\end{equation}

\textbf{Preconditioned gradient grafting.} To stabilize the optimization process of Eva-s, we adopt the idea of grafting from~\cite{anil2021scalable}, to take the direction of preconditioned gradient but use the step magnitude from the grafted optimizer such as SGD. Since maintaining a grafted optimizer requires extra memory consumption, we choose to graft Eva-s with the magnitude of the gradient, i.e., we layer-wisely scale the preconditioned gradient by a factor of $\sqrt{\grad_l^T\grad_l/\textbf{p}_l^T\textbf{p}_l}$.  

\subsection{Theoretical understanding}
\textbf{Trust regions of Eva-f and Eva-s.} Similar to Eva, Eva-f and Eva-s can be understood as trust-region optimization algorithms. For Eva-f, we have $\Delta \weight_l^T (I \otimes \bar{\ma}_{l-1} \bar{\ma}_{l-1}^T) \Delta \weight_l = \sum_{i=1}^{d_l} (\Delta \weight_{l,i}^T \bar{\ma}_{l-1})^2$, where $\weight_{l,i}$ is the $i$-th row of the parameter matrix $W_l$. In other words, it implies that the update step for each row of the parameter matrix $W_l$ is limited along the direction of $\bar{\ma}_{l-1}$. For Eva-s, we have $\Delta \weight_l^T (\mv_l\mv_l^T) \Delta \weight_l = (\Delta \weight_l^T\mv_l)^2$, which limits the layer-wise update size along the $\mv_l$ direction, as shown in Fig.~\ref{fig:trust-region}(b), where $\mv_l=\otimes_{i=1}^k \mv_{l,i}$. Consider the damping value for both Eva-f and Eva-s, the ball region is joined as well in the trust regions of Eva-f and Eva-s.

On the other hand, the relation between Eva-f and FOOF (Eva-s and Shampoo) is clear. Following Eq.~\ref{eq:A-a}, we have $I \otimes A_{l-1}A_{l-1}^T \succeq I \otimes \bar{\ma}_{l-1}\bar{\ma}_{l-1}^T$ ($\otimes_{i=1}^k M_{l,i} \succeq \otimes_{i=1}^k \mv_{l,i}\mv_{l,i}^T$), and it implies that FOOF (Shampoo) is a more conservative trust-region algorithm than Eva-f (Eva-s).

In summary, our Eva algorithms, including Eva-f and Eva-s, can be interpreted as more aggressive trust-region optimization methods than K-FAC, FOOF, and Shampoo, respectively. Thus, they share the fast convergence property of trust-region optimization in DNN training~\cite{Asi2019StochasticPP, Bae2022AmortizedPO}.

\begin{figure}[!ht]
    \centering
    \includegraphics[width=0.45\columnwidth]{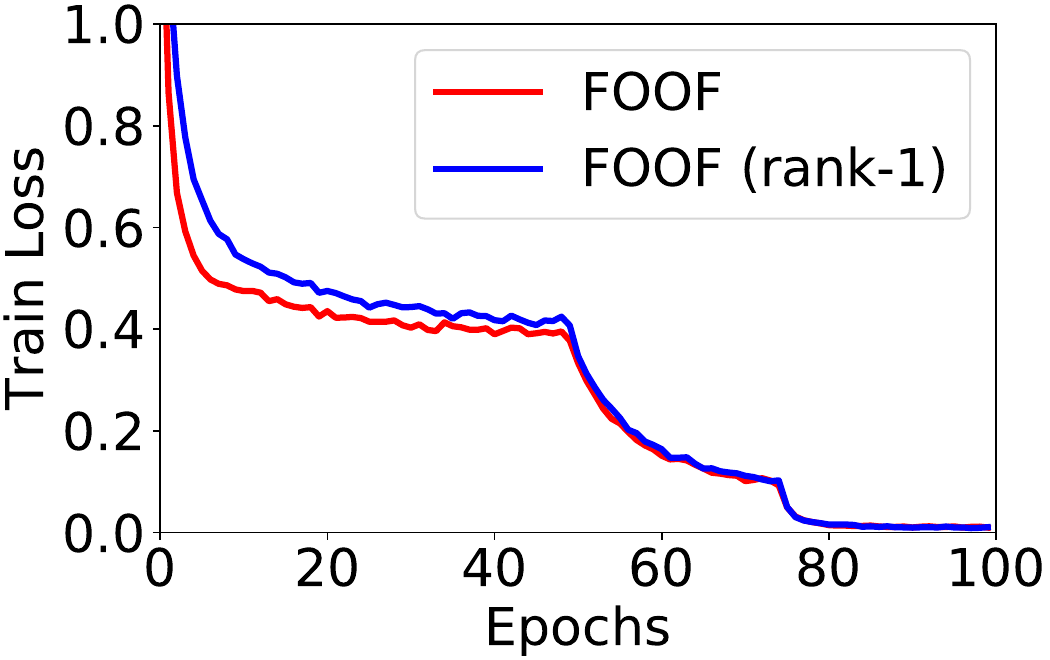}
    \includegraphics[width=0.45\columnwidth]{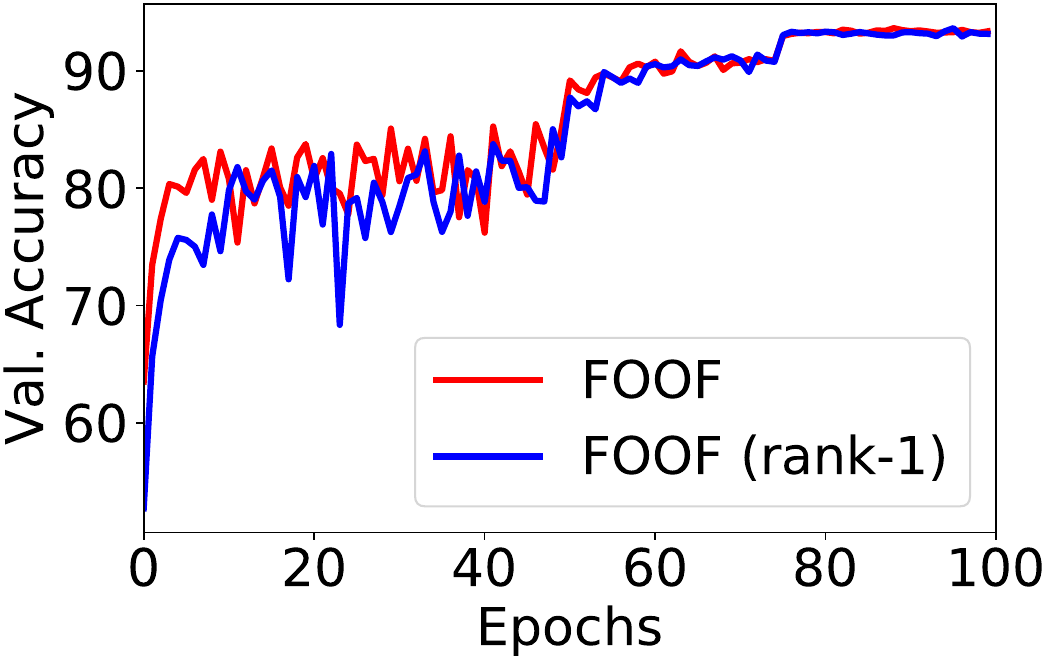}
    \caption{Convergence performance comparison between FOOF and FOOF (rank-1) algorithms for training VGG-19 on Cifar-10. }
    \label{fig:foof-rank1}
\end{figure}

\textbf{Eigen-decomposation on KFs.} 
Besides trust-region optimization, we provide two more observations for Eva-f. First, we find FOOF's KFs can be well approximated via the rank-1 eigen-decomposition. Formally, the update formula of FOOF can be approximated by
\begin{align}\label{eq:foof-approx}
    \Delta W_l &= -\alpha G_l [(R_{l} + \gamma I)^{-1} -\gamma^{-1}I + \gamma^{-1}I], \\
    &= -\alpha G_l [U((\Lambda+\gamma)^{-1}-\gamma^{-1}I)U^T + \gamma^{-1}I], \\
    &\approx -\frac{\alpha}{\gamma}(G_l - \frac{\lambda_1G_l\textbf{u}_1\mathbf{u}_1^T}{\gamma+\lambda_1}),  
\end{align}
where $R_l=U\Lambda U^T$, and $\lambda_1$ and $\textbf{u}_1$ are the largest eigen-value and corresponding eigen-vector. Since $R_l$ is typically low-rank (most $\lambda$s are zero), it can be well approximated by rank-1 eigen-decomposition, and gives a very similar update formula to Eva-f. In Fig.~\ref{fig:foof-rank1}, we validate that the approximate version of FOOF (rank-1) works as well as FOOF in the training loss and validation accuracy performance, which motivates us to vectorize FOOF with good performance guarantee. 

\begin{table*}[!ht]
    \centering
     \caption{Validation accuracy (\%) comparison between Eva and SGD/K-FAC algorithms for training from scratch with different epoch buckets. $\dagger$ indicates training with extra tricks. }
    \label{table:test-acc}
    \centering
    \begin{tabular}{cc|ccc|ccc}
    \hline
  \multirow{2}{*}{Model} & \multirow{2}{*}{Epoch} & \multicolumn{3}{c|}{Cifar-10} & \multicolumn{3}{c}{Cifar-100} \\ 
  ~ & ~ & SGD & K-FAC & Eva & SGD & K-FAC & Eva \\\hline 
  \multirow{3}{*}{VGG-19} & 50 & $90.95_{\pm 0.2}$ & $92.57_{\pm 0.3}$ & $\textbf{92.63}_{\pm 0.2}$ & $61.69_{\pm 0.8}$ & $70.14_{\pm 0.1}$ & $\textbf{70.23}_{\pm 0.7}$ \\ 
  ~ & 100 & $92.27_{\pm 0.3}$ & $\textbf{93.37}_{\pm 0.2}$ & $93.20_{\pm 0.1}$ & $67.97_{\pm 0.3}$ & $\textbf{72.25}_{\pm 0.1}$ & $71.79_{\pm 0.5}$ \\ 
  ~ & 200 & $93.02_{\pm 0.1}$ & $93.46_{\pm 0.2}$ & $\textbf{93.59}_{\pm 0.2}$ & $70.98_{\pm 0.0}$ & $\textbf{72.90}_{\pm 0.3}$ & $72.72_{\pm 0.4}$ \\\hline
  \multirow{3}{*}{ResNet-110} & 50 & $90.98_{\pm 0.6}$ & $\textbf{93.03}_{\pm 0.1}$ & $93.02_{\pm 0.3}$ & $67.93_{\pm 0.9}$ & $71.03_{\pm 0.5}$ & $\textbf{71.13}_{\pm 0.4}$ \\ 
  ~ & 100 & $92.49_{\pm 0.6}$ & $93.76_{\pm 0.1}$ & $\textbf{93.76}_{\pm 0.0}$ & $70.74_{\pm 0.7}$ & $72.31_{\pm 0.5}$ & $\textbf{72.38}_{\pm 0.2}$ \\
  ~ & 200 & $93.80_{\pm 0.2}$ & $\textbf{94.21}_{\pm 0.2}$ & $93.99_{\pm 0.1}$ & $72.43_{\pm 0.6}$ & $72.96_{\pm 0.3}$ & $\textbf{73.29}_{\pm 0.3}$ \\\hline
  \multirow{3}{*}{WRN-28-10$^\dagger$} & 50 & $95.28_{\pm 0.1}$ & $96.12_{\pm 0.2}$ & $\textbf{96.19}_{\pm 0.1}$ & $79.03_{\pm 0.1}$ & $80.98_{\pm 0.2}$ & $\textbf{81.15}_{\pm 0.2}$ \\ 
  ~ & 100 & $96.88_{\pm 0.2}$ & $\textbf{97.21}_{\pm 0.0}$ & $97.05_{\pm 0.0}$ & $83.03_{\pm 0.3}$ & $\textbf{83.54}_{\pm 0.4}$ & $83.52_{\pm 0.2}$ \\ 
  ~ & 200 & $97.33_{\pm 0.1}$ & $\textbf{97.44}_{\pm 0.1}$ & $97.38_{\pm 0.1}$ & $84.54_{\pm 0.0}$ & $\textbf{84.56}_{\pm 0.1}$ & $84.52_{\pm 0.1}$ \\ \hline
    \end{tabular}
\end{table*}

\textbf{Gradient descent on neurons. } 
Second, as Eva-f is a vectorized version of FOOF, it is of interest to establish the link to gradient descent on neurons as in~\cite{Benzing2022Foof}. That is, Eva-f can be regarded as finding a weight update $\Delta W_l$ so that the layer's \textit{expected} output changes in the gradient direction, formally, 
\begin{align} \label{eq:sgd-neuron}
    & \min_{\Delta W_l} \sum_{i=1}^{n} \frac{1}{n} ||\Delta W_l \bar{\ma}_{l-1} - \mb_l^{(i)}||^2, \\
    = & \min_{\Delta W_l} ||\Delta W_l \bar{\ma}_{l-1} \bar{\ma}_{l-1}^T - G_l ||^2, 
\end{align}
where $\Delta W_l \bar{\ma}_{l-1}$ is the expected output change at layer $l$, and $\mb_l^{(i)}$ is the pre-activation gradient w.r.t. $W_l\ma_{l-1}^{(i)}$ on the $i$-th sample of a mini-batch. This problem has a good solution, i.e., $\Delta W_l=G_l(\bar{\ma}_{l-1}\bar{\ma}_{l-1}^T+\gamma I)^{-1}$ , which results in a small loss of $\gamma^2 ||\Delta W_l||^2$ and gives exactly the preconditioned gradient of Eva-f. To derive FOOF~\cite{Benzing2022Foof}, the major difference is to use individual input $\ma_{l-1}^{(i)}$ instead of expected input $\bar{\ma}_{l-1}$ in the objective Eq.~\ref{eq:sgd-neuron}. 

\section{Evaluation}

\subsection{Convergence validation}
We follow the common configurations of prior second-order optimization methods~\cite{martens2015optimizing, Goldfarb20quasi, Ren21TNT}, and validate the optimization performance of our Eva algorithm by training an 8-layer autoencoder (with hidden dimensions of $[1000, 500, 250, 30, 250, 500, 1000]$) on four small datasets: MNIST, FMIST, FACES, and CURVES. 

\begin{figure}[!ht]
    \centering
    \subfloat[MNIST]{
        \includegraphics[width=0.45\columnwidth]{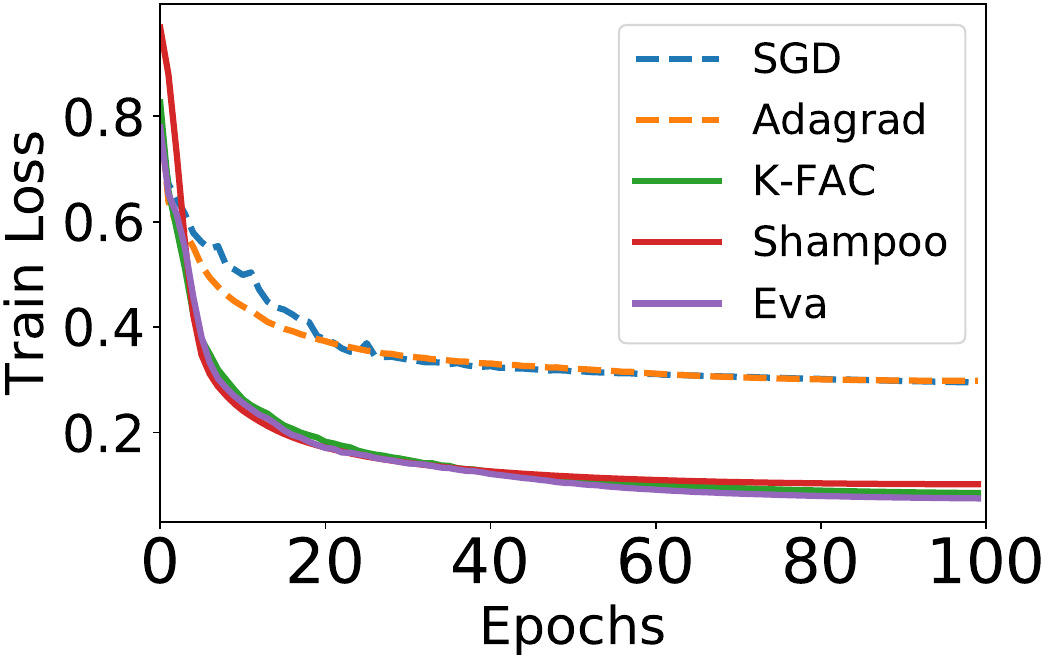}
    }
    \subfloat[FMNIST]{
        \includegraphics[width=0.45\columnwidth]{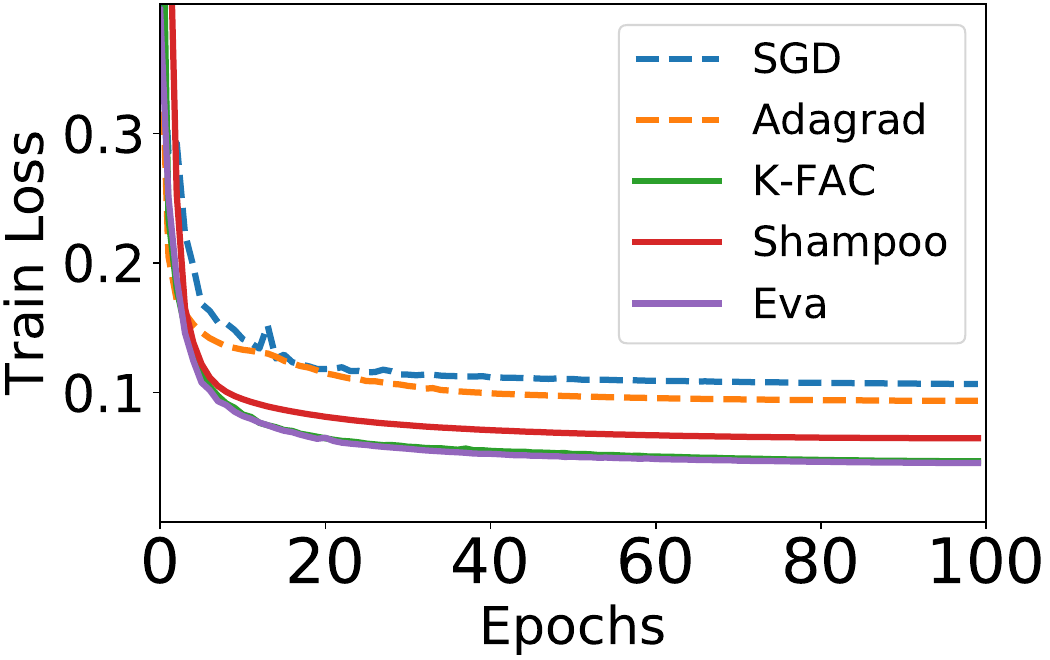}
    } \\
    \subfloat[FACES]{
        \includegraphics[width=0.45\columnwidth]{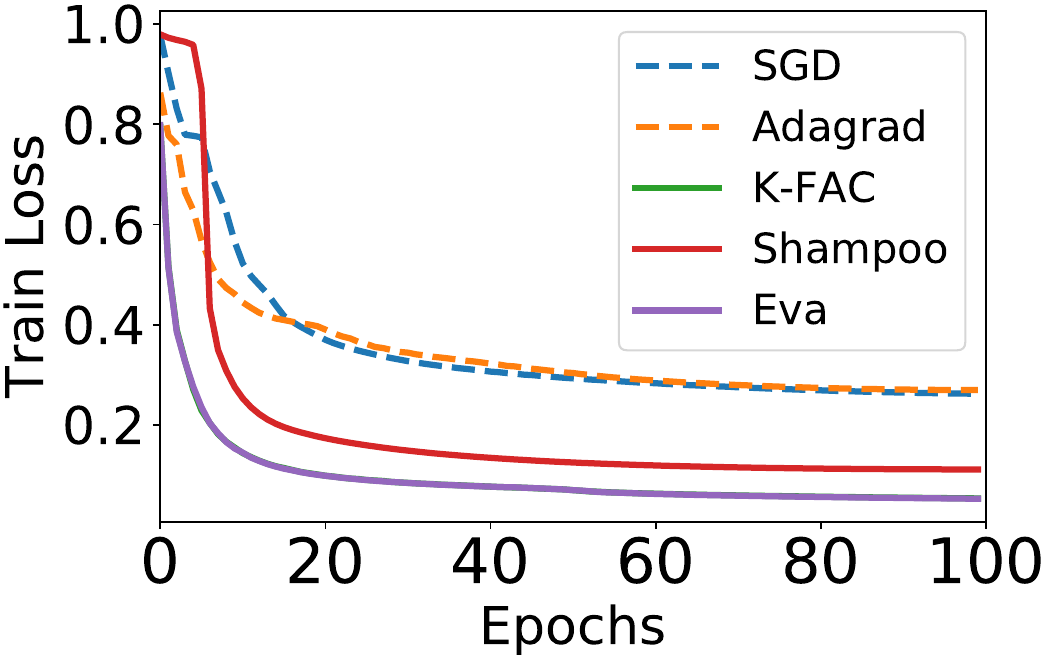}
    } 
    \subfloat[CURVES]{
        \includegraphics[width=0.45\columnwidth]{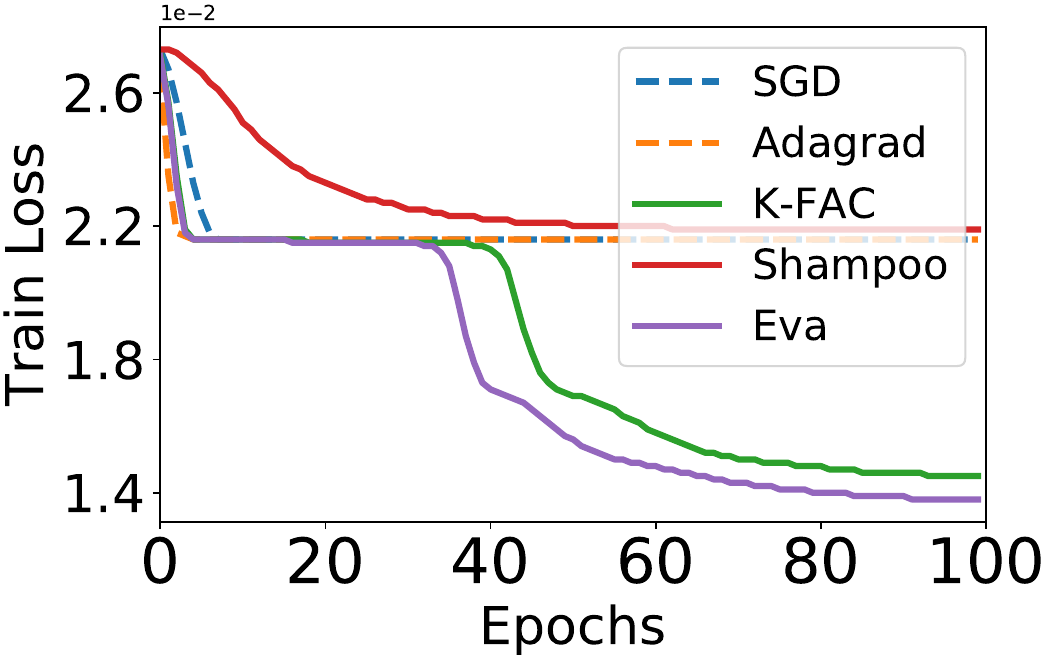}
    }
    \caption{Optimizing an 8-layer autoencoder on MNIST, FMIST, FACES, and CURVES datasets with different algorithms.}
    \label{fig:mnist-ae}
\end{figure}

We set batch size to 1000, and run each algorithm for 100 epochs with a linear decay learning rate schedule. We tune the learning rate in the range of $[0.001, 0.5]$ for different cases. We choose first-order methods SGD (with a momentum of 0.9) and Adagrad~\cite{Duchi2010Adagrad}, and second-order methods K-FAC~\cite{martens2015optimizing} and Shampoo~\cite{Gupta2018ShampooPS} as our baselines. The results are given in Fig.~\ref{fig:mnist-ae}. 

The Fig.~\ref{fig:mnist-ae} shows that second-order methods K-FAC and Eva optimize much faster than SGD in this task, and our Eva can optimize the autoencoder model at the same convergence speed as K-FAC. In addition, one can see that Eva performs closely or better than another second-order method Shampoo, and Shampoo is faster than Adagrad except its unstable result on CURVES. Note that Shampoo is a full-matrix version of Adagrad. We also compare the validation loss performance of each algorithm, and the results are very close to those of optimization comparison, that is, Eva performs similarly to K-FAC and outperforms other first-order counterparts on four datasets. 

\begin{table*}[!ht]
    \centering
     \caption{Relative iteration time and memory over SGD. Values in parentheses represent the results with increased second-order update intervals (10 on Cifar-10 and 50 on ImageNet).}
    \label{table:efficiency}
    \centering
    \begin{tabular}{cc|cc|cc|cc}
    \hline
  \multirow{2}{*}{Dataset} & \multirow{2}{*}{Model} & \multicolumn{2}{c|}{Shampoo} & \multicolumn{2}{c|}{K-FAC} & \multicolumn{2}{c}{Eva} \\ 
  ~ & ~ & Time & Mem & Time & Mem & Time & Mem \\ \hline 
  \multirow{3}{*}{Cifar-10} & VGG-19    & 19.6$\times$ (2.89$\times$) & 1.01$\times$ & 5.57$\times$ (1.68$\times$) & 1.01$\times$ & \textbf{1.13}$\times$ & \textbf{1.00}$\times$ \\ 
  & ResNet-110   & 6.79$\times$ (1.90$\times$) & 1.00$\times$ & 1.64$\times$ (1.16$\times$) & 1.03$\times$ & \textbf{1.16}$\times$ & \textbf{1.00}$\times$ \\ 
  & WRN-28-10 & 6.69$\times$ (1.60$\times$) & 1.05$\times$ & 2.68$\times$ (1.25$\times$) & 1.38$\times$ & \textbf{1.03}$\times$ & \textbf{1.00}$\times$ \\ \hline
  \multirow{3}{*}{ImageNet} & ResNet-50    & 30.7$\times$ (1.71$\times$) & 1.07$\times$ & 2.52$\times$ (1.14$\times$) & 1.06$\times$ & \textbf{1.09}$\times$ & \textbf{1.00}$\times$ \\ 
  & Inception-v4 & 75.3$\times$ (2.70$\times$) & 1.11$\times$ & 3.95$\times$ (1.28$\times$) & 1.42$\times$ & \textbf{1.28}$\times$ & \textbf{1.00}$\times$ \\ 
  & ViT-B/16     & 199.$\times$ (6.20$\times$) & 1.31$\times$ & 4.47$\times$ (1.43$\times$) & 1.45$\times$ & \textbf{1.18}$\times$ & \textbf{1.00}$\times$ \\ \hline
  \end{tabular}
\end{table*}
 
\subsection{Generalization performance}
Next, we evaluate the generalization performance of Eva in real-world image classification applications with three representative models, VGG-19~\cite{simonyan2014very}, ResNet-110~\cite{he2016deep}, and WideResNet-28-10 (WRN-28-10)~\cite{wide-resnet} on Cifar-10~\cite{krizhevsky2009learning} and Cifar-100~\cite{krizhevsky2009learning} datasets. We compare Eva to the first-order baseline SGD (with a momentum of $0.9$), and the second-order baseline K-FAC~\cite{martens2015optimizing}. Following the configurations of~\cite{pauloski2020convolutional,pauloski2021kaisa}, we set the same hyper-parameters for all algorithms for a fair comparison, and the details are given in Appendix~C.1 in~\cite{lin2022eva}. Since training with more epochs can generalize better~\cite{hoffer2017train}, we run each algorithm with $50$, $100$, and $200$ epochs for a better comparison from compressed to sufficient training budgets. The validation accuracy comparison is shown in Table~\ref{table:test-acc} and Table~\ref{table:test-acc-shampoo}, where we report the mean and std over three independent runs. We summarize the results in the following three aspects. 

First, in the models VGG-19 and ResNet-110 with commonly used settings from their original papers (details in Appendix~C.1 in~\cite{lin2022eva}), it is seen that Eva performs closely to the second-order K-FAC, and they both consistently outperform SGD under different training budgets. To be specific, under the 50-epoch setting, both Eva and K-FAC outperform SGD by a large margin, e.g., $+8.5\%$ for training VGG-19 on Cifar-100; under the setting of sufficient 200 epochs, Eva and K-FAC still achieve slightly better generalization performance than SGD. Eva and K-FAC with 50 (and 100) epochs achieve the loss or validation accuracy that SGD needs to take 100 (and 200) epochs. In summary, our proposed second-order Eva has similar good convergence performance with K-FAC, and both of them train the models around $2\times$ faster than SGD in terms of iterations. It validates that second-order algorithms have better convergence performance than SGD and they can reach the same target accuracy in fewer number of training iterations. 

Second, considering that existing training paradigms typically use extra tricks like CutMix~\cite{Yun2019CutMix} and AutoAugment~\cite{Cubuk2019AutoAugment} to achieve better validation accuracy, we train a relatively new model WRN-28-10~\cite{wide-resnet} on Cifar-10 and Cifar-100 to compare the performance of different optimizers. Note CutMix~\cite{Yun2019CutMix} and AutoAugment~\cite{Cubuk2019AutoAugment} have been particularly developed and heavily tuned for the first-order SGD optimizer, which could be detrimental to second-order algorithms. However, as shown in Table~\ref{table:test-acc}, our Eva and K-FAC still learn faster and generalize better than SGD under the same number of training epochs. Additional results to verify the generalization performance for fine-tuning pretrained models are provided in Table~\ref{table:test-acc-finetuning}, showing that second-order optimization methods such as K-FAC and Eva can generalize as well as SGD on finetuning pretrained models, even though they were pretrained with first-order algorithms (SGD for EfficientNet-b0, and AdamW for ViT-B/16). In summary, Eva achieves the same generalization performance with a less number of iterations than well-tuned SGD or it achieves higher generalization performance with the same training number of iterations as SGD. 

\begin{table}[!ht]
    \centering
     \caption{Validation accuracy (\%) comparison between Eva and SGD/K-FAC algorithms for finetuning pretrained models with 20 epochs. EffNet stands for EfficientNet. }
    \label{table:test-acc-finetuning}
    \centering
    \begin{tabular}{cc|ccc}
    \hline
    Dataset & Model & SGD & K-FAC & Eva \\ \hline
    \multirow{2}{*}{Cifar-10} & EffNet-b0 & $97.39_{\pm 0.1}$ & $97.37_{\pm 0.0}$ & $\textbf{97.43}_{\pm 0.1}$ \\
     & ViT-B/16 & $98.87_{\pm 0.0}$ & $98.87_{\pm 0.0}$ & $\textbf{98.88}_{\pm 0.0}$ \\ \hline
    \multirow{2}{*}{Cifar-100} & EffNet-b0 & $\textbf{85.41}_{\pm 0.0}$ & $85.32_{\pm 0.2}$ & $85.38_{\pm 0.0}$  \\
    & ViT-B/16 & $92.79_{\pm 0.1}$ & $92.68_{\pm 0.2}$ & $\textbf{92.88}_{\pm 0.0}$ \\ \hline
    \end{tabular}
\end{table}

\begin{table}[!ht]
    \centering
    \caption{Validation accuracy (\%) comparison between Eva and 4 more algorithms for training from scratch on Cifar-10 with 100 epochs. $\dagger$ indicates training with extra tricks (except M-FAC).} 
    \label{table:test-acc-shampoo}
    \centering
    \begin{tabular}{ccccc}
    \hline
    Model & Adagrad & AdamW & Shampoo & M-FAC \\ \hline
    VGG-19 & $92.42_{\pm 0.0}$ & $92.97_{\pm 0.1}$ & $\textbf{93.38}_{\pm 0.1}$ & $92.50_{\pm 0.1}$ \\
    ResNet-110 & $90.34_{\pm 0.2}$ & $92.61_{\pm 0.1}$ & $92.47_{\pm 0.2}$ & $\textbf{93.45}_{\pm 0.1}$ \\ 
    WRN-28-10$^\dagger$ & $93.72_{\pm 0.2}$ & $96.91_{\pm 0.0}$ & $\textbf{96.99}_{\pm 0.1}$ & $94.54_{\pm 0.2}$ \\ \hline
    \end{tabular}
\end{table}

Third, we compare the convergence performance of Eva with 4 more popular optimizers: Adagrad~\cite{Duchi2010Adagrad}, AdamW~\cite{Loshchilov2019adamw}, Shampoo~\cite{Gupta2018ShampooPS}, and M-FAC~\cite{mfac2021frantar}. Adagrad and AdamW are common adaptive gradient methods, while Shampoo and M-FAC are recently proposed second-order algorithms. We tune the learning rate for each algorithm to choose a best one for particular algorithms to train three models on Cifar-10 for 100 epochs. The results are given in Table~\ref{table:test-acc-shampoo}, showing that Eva achieves comparable performance to other second-order algorithms Shampoo and M-FAC, and outperforms first-order adaptive methods Adagrad and AdamW on different models. M-FAC, however, results in $2.5\%$ accuracy loss, compared to Eva and Shampoo in training WRN-28-10. Note that we exclude CutMix and AutoAugment in M-FAC as they cause M-FAC divergence in training WRN-28-10. 

\subsection{Time and memory efficiency}
As we have shown the good convergence and generalization performance of Eva, we demonstrate its per-iteration training time and memory consumption compared with SGD, Shampoo, and K-FAC. We do not report M-FAC here as it needs to store extra $m$ gradients for FIM estimation ($m=1024$ by default), which is very memory inefficient (normally causes out-of-memory). To cover relatively large models, we select three extra popular deep models trained on ImageNet~\cite{deng2009imagenet}: ResNet-50, Inception-v4~\cite{szegedy2017inception}, and ViT-B/16~\cite{alexey21vit}. We run each algorithm on an Nvidia RTX2080Ti GPU. Following the training recipes in~\cite{pauloski2020convolutional}, we set the second-order update interval as 10 on Cifar-10, and 50 on ImageNet for both Shampoo and K-FAC to reduce their average iteration time. More detailed configurations can be found in Appendix~C.1 in~\cite{lin2022eva}. We report relative time and memory costs over SGD in Table~\ref{table:efficiency}. 

\textbf{Time efficiency.} We can see that Eva has the shortest iteration time among all evaluated second-order algorithms. Eva has only $1.14\times$ longer iteration time on average than SGD, but it achieves an average of $3.04\times$ and $49.3\times$ faster than K-FAC and Shampoo, respectively. This is because our Eva does not need to explicitly compute the inverse of second-order matrix while K-FAC requires expensive inverse matrix computations on KFs and Shampoo performs even more computations of inverse $p$-th roots ($p \geq 2$). Therefore, Eva is able to update second-order information iteratively to achieve faster convergence, but K-FAC and Shampoo have to update their preconditioners infrequently. For example, the average time can be reduced to $1.32\times$ and $2.83\times$ for K-FAC and Shampoo, respectively, when increasing the second-order update interval (10 on Cifar-10 and 50 on ImageNet). We will show the end-to-end training performance in \S\ref{subsec:exp-endtoend}.

\textbf{Memory efficiency.} In terms of memory consumption, Eva takes almost the same memory as SGD, and spends much less memory than Shampoo (with $1.09\times$) and K-FAC (with $1.22\times$), since Eva only needs to store small vectors, but Shampoo and K-FAC have to store second-order matrices and their inverse results. We notice that K-FAC consumes more GPU memory than Shampoo as K-FAC requires extra memory on intermediate states to estimate KFs, such as unfolding images. 

In summary, Eva is a much more efficient second-order optimizer than K-FAC and Shampoo in terms of the average iteration time and memory consumption. 

\subsection{End-to-end training performance}\label{subsec:exp-endtoend}
We further compare the end-to-end wall-clock time to reach the target accuracy training with different optimizers. We train VGG-19, ResNet-110, and WRN-28-10 on Cifar-10 with one RTX2080Ti GPU (11GB memory), and ResNet-50 on ImageNet with 32 RTX2080Ti GPUs. We set the second-order information update interval to 10 on Cifar-10 and 50 on ImageNet for K-FAC and Shampoo. More configurations can be found in Appendix~C.1 in~\cite{lin2022eva}. 

\begin{figure}[!ht]
    \centering
    \subfloat[VGG-19, Cifar-10]{
        \includegraphics[width=0.45\columnwidth]{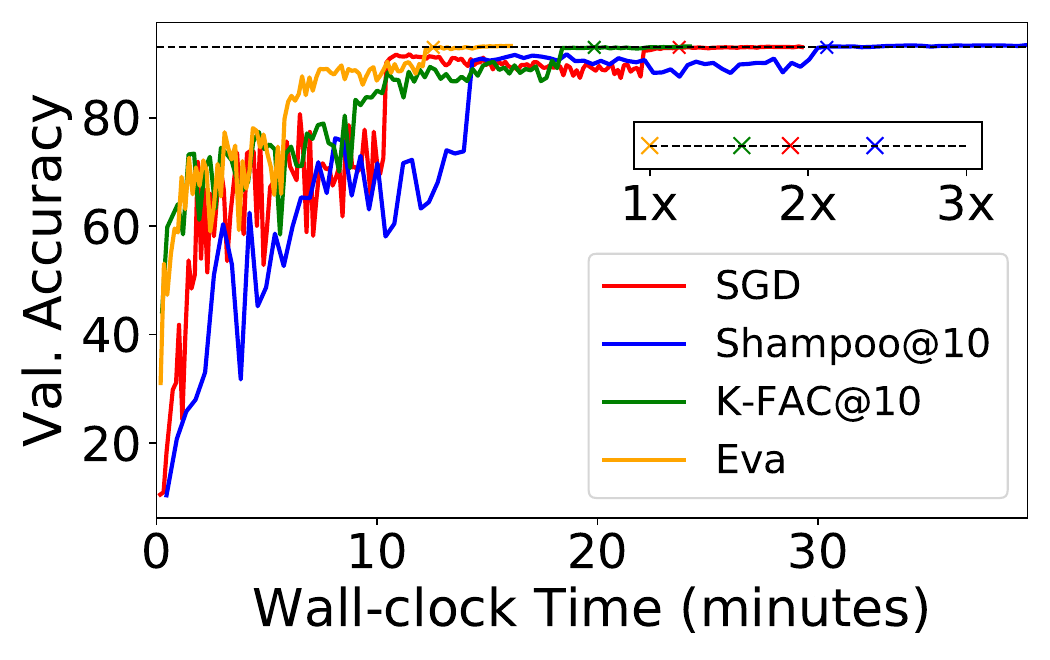}
    }
    \subfloat[ResNet-110, Cifar-10]{
        \includegraphics[width=0.45\columnwidth]{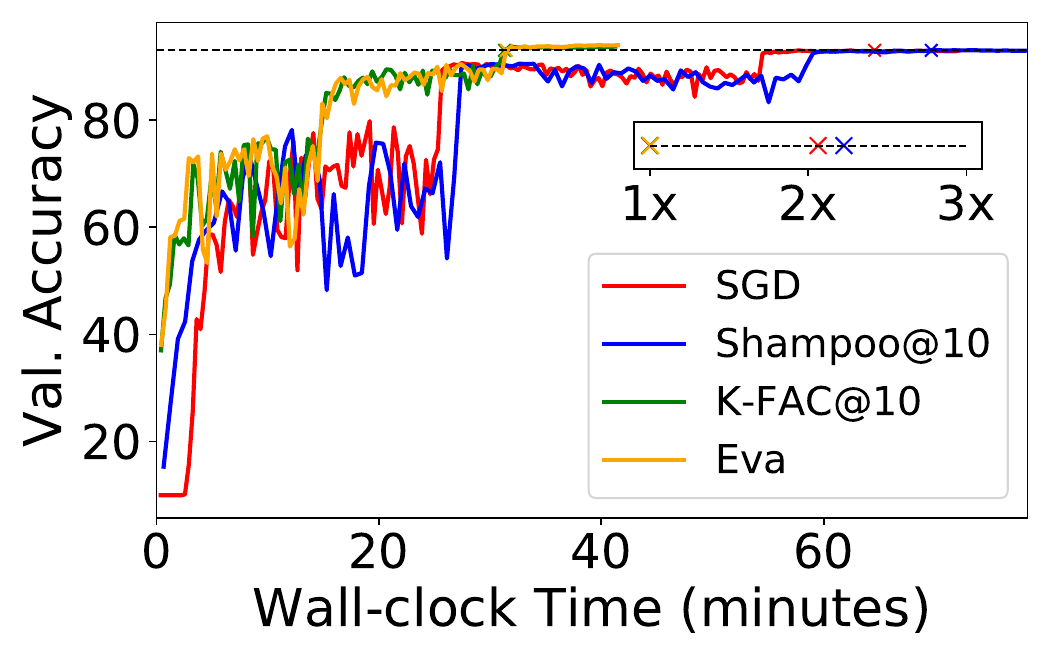}
    }\\
    \subfloat[WRN-28-10, Cifar-10]{
        \includegraphics[width=0.45\columnwidth]{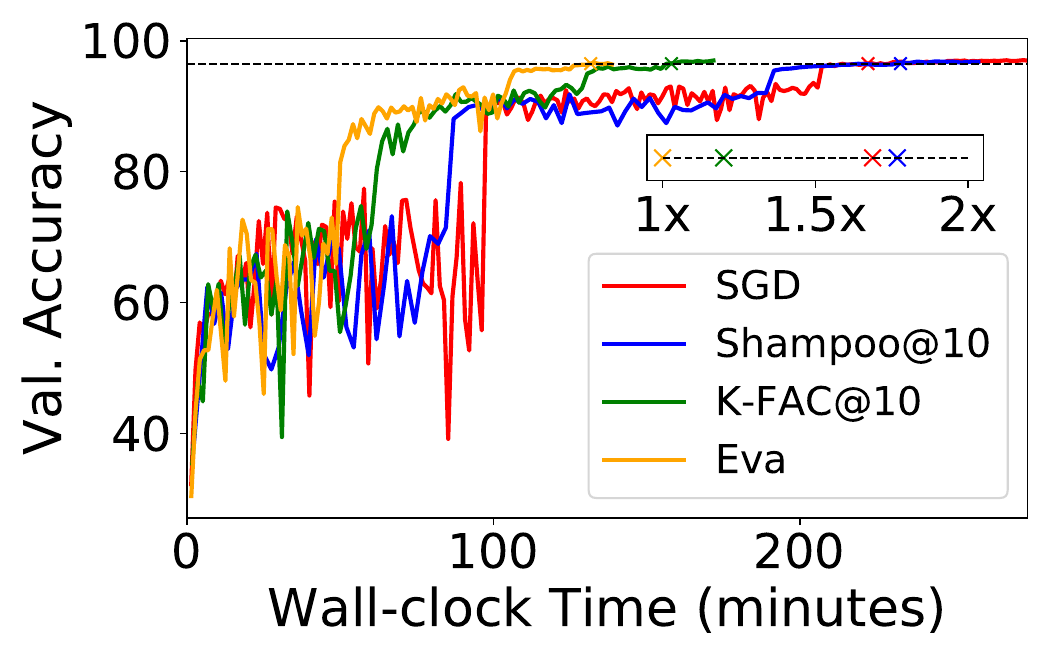}
    }
    \subfloat[ResNet-50, ImageNet]{
        \includegraphics[width=0.45\columnwidth]{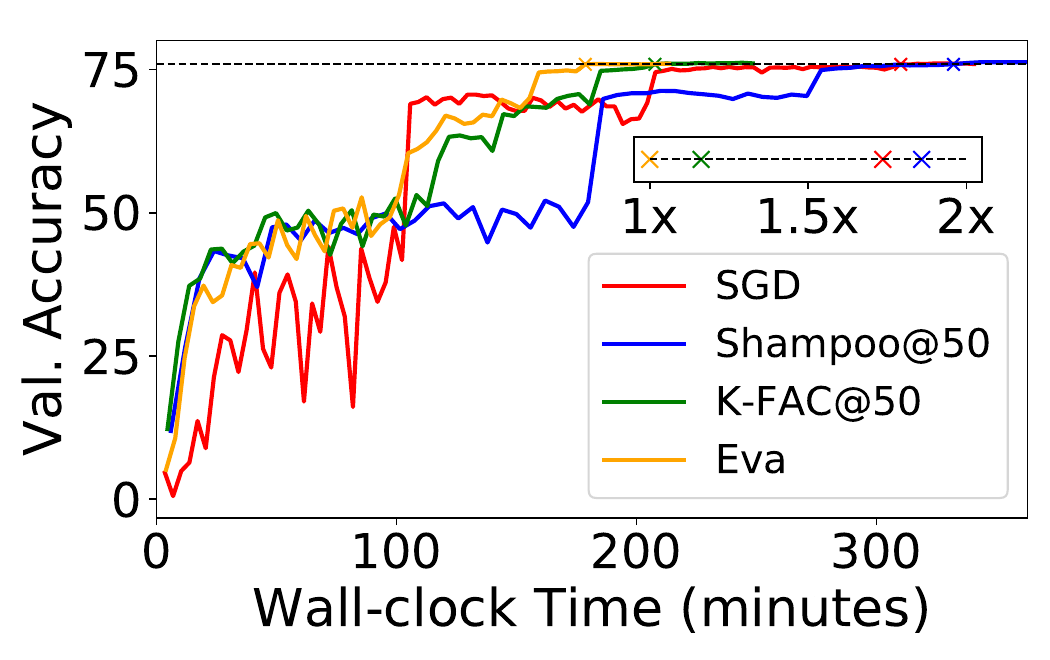}
    }
    \caption{Wall-clock time comparison between Eva and SGD/K-FAC/Shampoo algorithms for training multiple DNNs. The inset plot reports relative time-to-solution over Eva. }
    \label{fig:end2end}
\end{figure}
\begin{figure}[!ht]
    \centering
    \subfloat[ResNet-110]{
        \includegraphics[width=0.45\columnwidth]{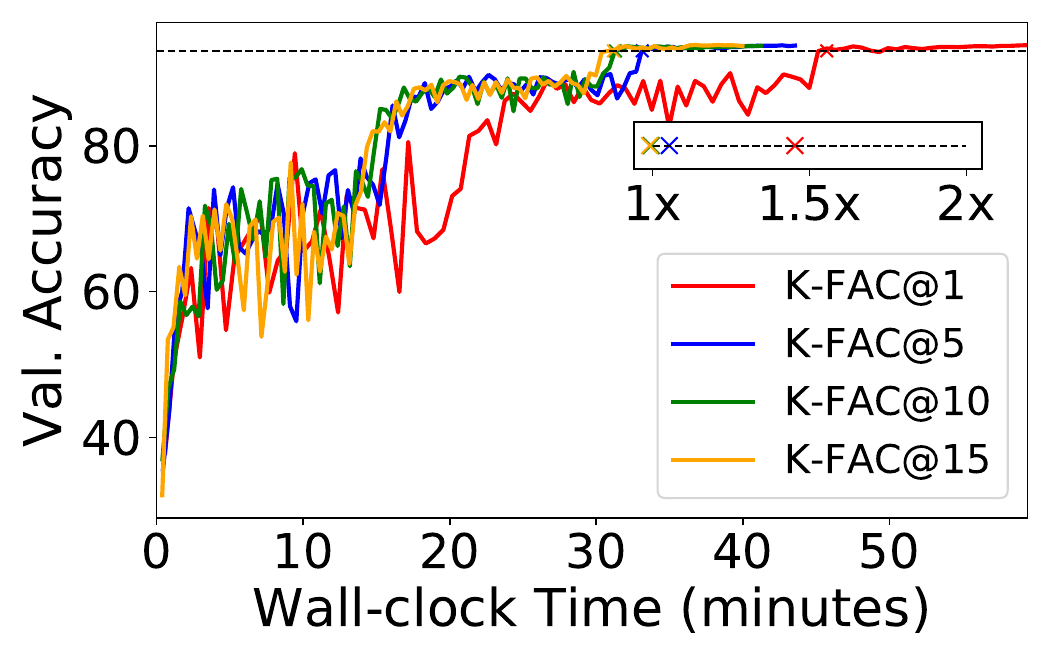}
    }
    \subfloat[VGG-19]{
        \includegraphics[width=0.45\columnwidth]{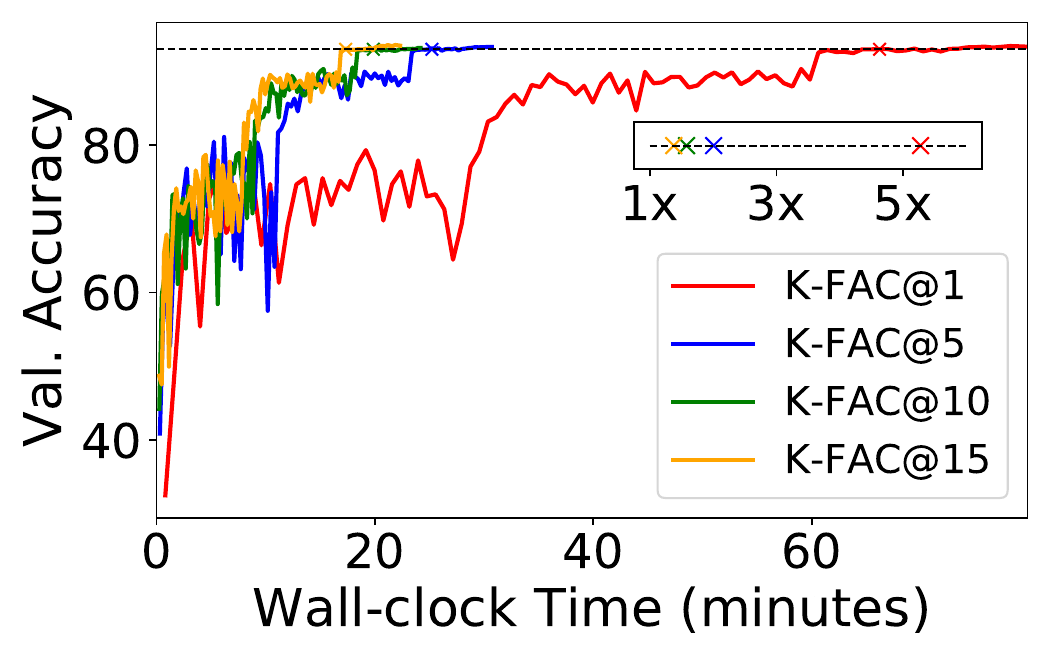}
    }
    \caption{Wall-clock time comparison of K-FAC with different update intervals on Cifar-10. The inset plot reports relative time-to-solution over Eva. }
    \label{fig:convergence-kfac-freq}
\end{figure}

The results are given in Fig.~\ref{fig:end2end}, showing that Eva optimizes faster than all the other evaluated algorithms SGD, K-FAC, and Shampoo. On the Cifar-10 dataset, Eva is $1.88\times$, $1.26\times$, and $2.14\times$ faster on average than SGD, K-FAC, and Shampoo, respectively. This is because Eva requires less training epochs than SGD for convergence, and the iteration time of Eva is smaller than K-FAC and Shampoo. We notice that K-FAC is also possible to optimize faster than SGD, but it needs to increase the second-order update interval. Otherwise, the training time of K-FAC would be much more expensive as studied in Fig.~\ref{fig:convergence-kfac-freq}. 

In Fig.~\ref{fig:convergence-kfac-freq}, K-FAC@1 takes much longer training time than Eva, i.e., $1.45\times$ and $5.27\times$ than Eva for ResNet-110 and VGG-19, respectively. Thus, it is often required to increase the second-order update interval to make K-FAC more affordable. For example, the training speed of K-FAC@10 (update KFs and their inverses every 10 iterations) is comparable to Eva for training ResNet-100, and it is $1.58\times$ slower than Eva for training VGG-19 to achieve the target accuracy. Unlike K-FAC, our Eva can converge quickly using an update interval of second-order information of 1, i.e., updating second-order information iteratively, which avoids the efforts of tuning the interval and the danger of performance degradation with stale information. 

Besides, while increasing the interval makes K-FAC and Shampoo computationally efficient, they require more GPU memory than Eva. For a fair comparison, on the large-scale ImageNet dataset, we set per-GPU batch size to 96 for SGD and Eva, and 64 for K-FAC and Shampoo, to maximize the GPU utilization. In this setting, the results shown in Fig.~\ref{fig:end2end}(d) indicates that Eva achieves $1.74\times$, 	$1.16\times$, $1.86\times$ speedups over SGD, K-FAC, and Shampoo, respectively, to achieve the target accuracy of 75.9\% on the validation set according to MLPerf. The median test accuracy of the final 5 epochs on
ImageNet is 76.02\%, 76.25\%, 76.06\%, and 75.96\% for SGD, Shampoo, K-FAC, and Eva, with 100, 60, 55, and 55 epochs, respectively in training ResNet-50. The throughput improvement of Eva by using larger per-GPU batch size is studied in Table~\ref{table:batch-size-throughput}, where we set the per-GPU batch size to maximally utilize GPU memory, i.e., 64 for K-FAC@50 and Shampoo@50, and 96 for SGD and Eva. 
\begin{table}[!ht]
    \centering
     \caption{Throughput (number of samples per second) comparison on ImageNet.}
    \label{table:batch-size-throughput}
    \centering
    \begin{tabular}{c|cccc}
    \hline
    Algorithm  & SGD & Eva & Shampoo@50 & K-FAC@50 \\\hline
    Batch Size & 96 & 96 & 64 & 64  \\
    Throughput & 7420.3 & 6857.1 & 4366.7 & 5520.2  \\\hline
    \end{tabular}
\end{table}

\subsection{Hyper-parameter and ablation study} \label{sec:hyper}
\begin{figure}[!ht]
    \centering
    \subfloat[Learning rate]{
        \includegraphics[width=0.45\columnwidth]{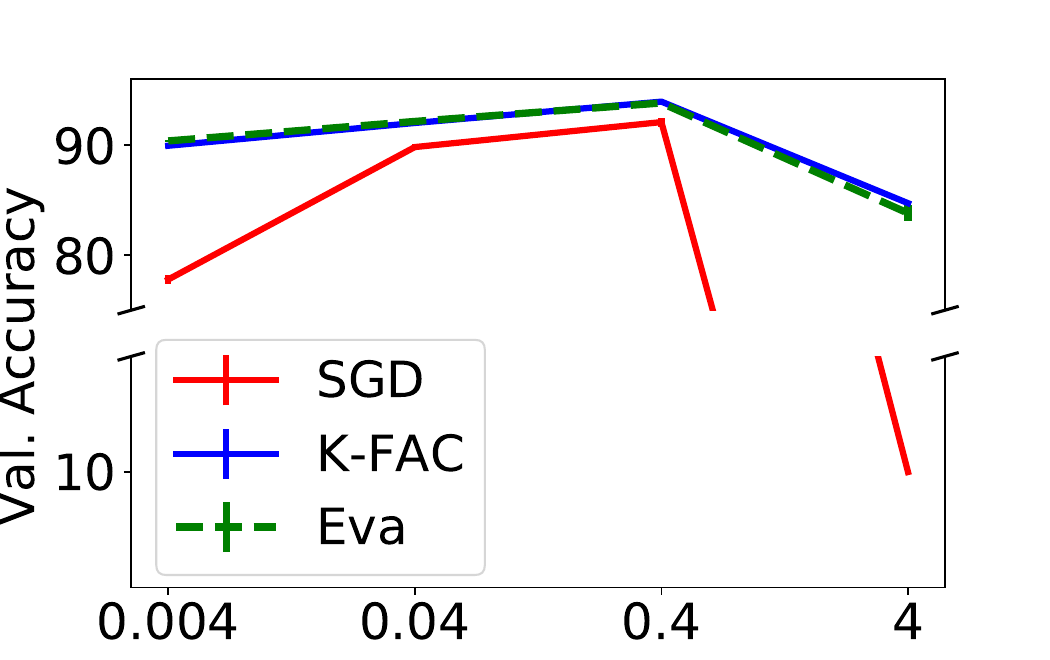}
    }
    \subfloat[Batch size]{
        \includegraphics[width=0.45\columnwidth]{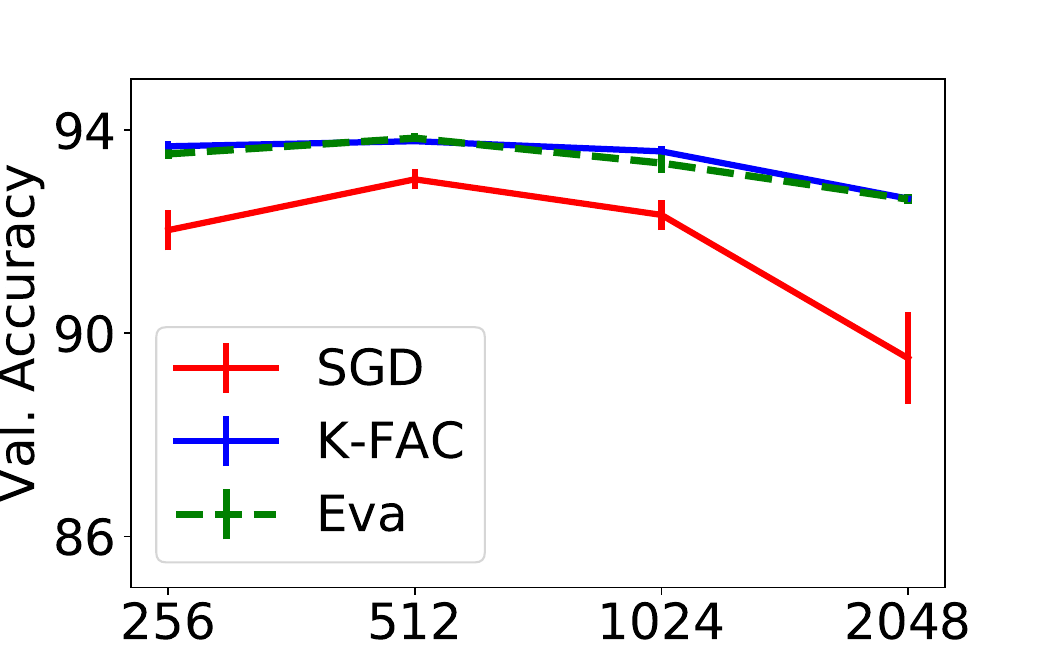}
    }\\
    \subfloat[Damping]{
        \includegraphics[width=0.45\columnwidth]{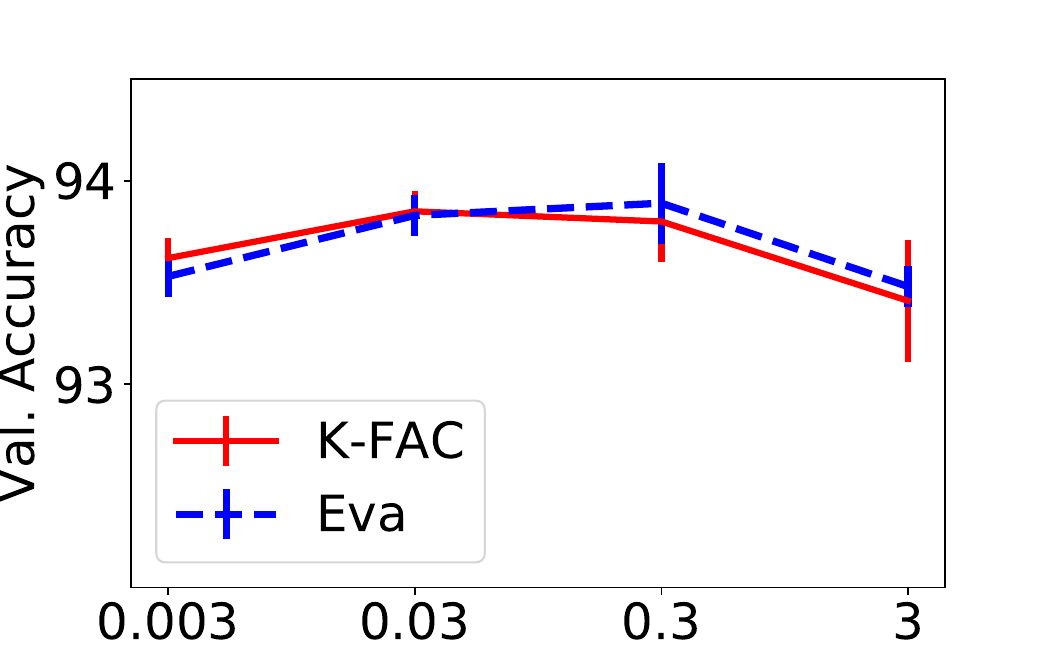}
    }
    \subfloat[Running average]{
        \includegraphics[width=0.45\columnwidth]{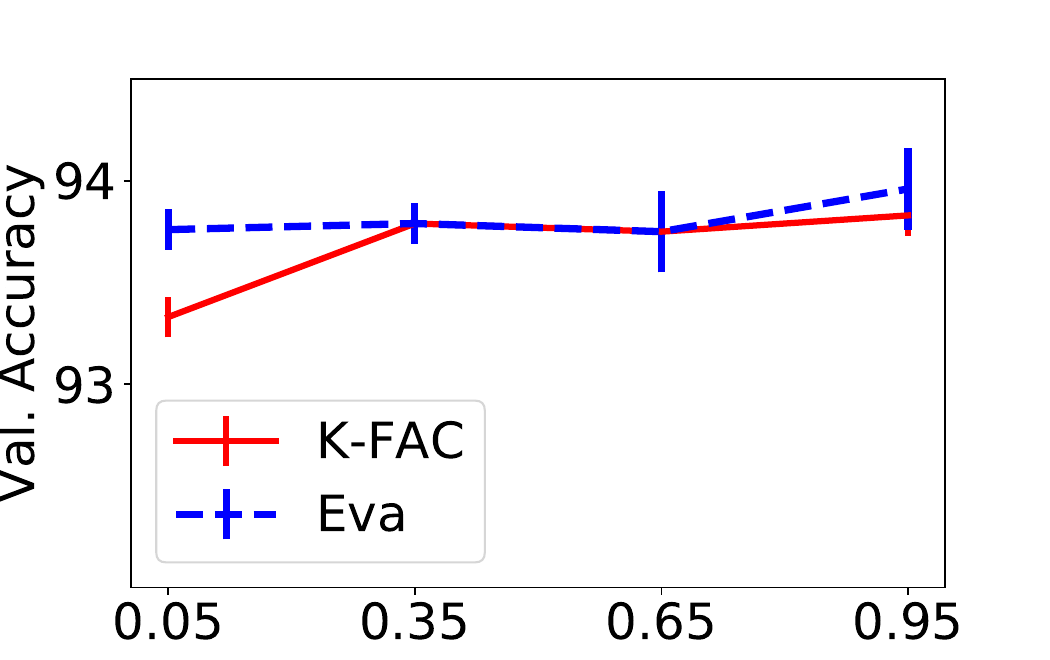}
    }
    \caption{Hyper-parameter study of Eva by training ResNet-110 on Cifar-10 with 100 epochs. }
    \label{fig:convergence-hyper}
\end{figure}

We study the convergence performance of Eva with different hyper-parameters, including learning rate, batch size, damping, and running average. The results are shown in Fig.~\ref{fig:convergence-hyper} in training ResNet-110 on Cifar-10, and the similar results of training VGG-19 on Cifar-100 are reported in Appendix~C.2 in~\cite{lin2022eva}. 

First, the learning rate and batch size are two key hyper-parameters to SGD, K-FAC, and Eva, as shown in Fig.~\ref{fig:convergence-hyper}(a) and (b), but Eva and K-FAC can consistently outperform SGD under different settings, and largely outperform SGD in the large batch size regime. However, a large learning rate would cause performance degradation in SGD and Eva, but Eva is more robust to the learning rate and batch size than SGD. Second, we tune the damping and running average hyper-parameters introduced in K-FAC and Eva, as shown in Fig.~\ref{fig:convergence-hyper}(c) and (d). K-FAC and Eva perform very closely and they are both robust to damping and running average values, which implies that one can use default hyper-parameters to achieve good performance. Therefore, the hyper-parameters used in SGD and some dedicated hyper-parameters used in K-FAC can be directly applied in Eva to achieve good convergence performance instead of tuning them.

We also conduct the ablation study on Eva to validate the necessity of using momentum, KL clip, and KVs. We compare the performance of Eva to three variants without using momentum, KL clip (see Eq.~\ref{eq:kl-clip}), and KVs (see Appendix~B.2 in~\cite{lin2022eva}), respectively. The results are demonstrated in Table~\ref{table:ablation}, which show that discarding any part of them would cause performance degradation. Specifically, momentum can improve the generalization of Eva (similar to SGD with momentum). KL clip is of importance to prevent exploding the preconditioned gradients (similar to K-FAC), otherwise, Eva w/o KL clip could cause divergence as shown in training WRN-28-10 on Cifar-10. In addition, KVs are required, rather than gradient norm, to construct useful curvature information that helps optimization. 

\begin{table}[!ht]
    \centering
     \caption{Ablation study on Eva without using momentum (m.), KL clip, and KVs, respectively. We train ResNet-110 and WRN-28-10$^\dagger$ on Cifar-10, and VGG-19 on Cifar-100. } 
    \label{table:ablation}
    \centering
    \addtolength{\tabcolsep}{-1.2pt}
    \begin{tabular}{ccccc}
    \hline
    Model & Eva & w/o m. & w/o KL clip & w/o KVs \\ \hline
    ResNet-110 & $\textbf{93.86}_{\pm 0.1}$ & $89.39_{\pm 0.1}$ & $90.95_{\pm 0.5}$ & $92.62_{\pm 0.2}$ \\
    WRN-28-10$^\dagger$ & $\textbf{97.03}_{\pm 0.1}$ & $94.59_{\pm 0.1}$ & $67.27_{\pm 50.}$ & $96.67_{\pm 0.0}$ \\
    VGG-19     & $\textbf{72.04}_{\pm 0.3}$ & $66.00_{\pm 0.3}$ & $60.93_{\pm 2.3}$ & $66.64_{\pm 0.8}$ \\ \hline
    \end{tabular}
\end{table}

\subsection{Extensions to Eva-f and Eva-s}
We evaluate two more vectorized approximation algorithms: Eva-f and Eva-s, where Eva-f and Eva-s are vectorized FOOF~\cite{Benzing2022Foof} and Shampoo~\cite{Gupta2018ShampooPS}, respectively. 

\begin{figure}[!t]
    \centering
    \subfloat[Eva-f, Cifar-10]{
        \includegraphics[width=0.45\columnwidth]{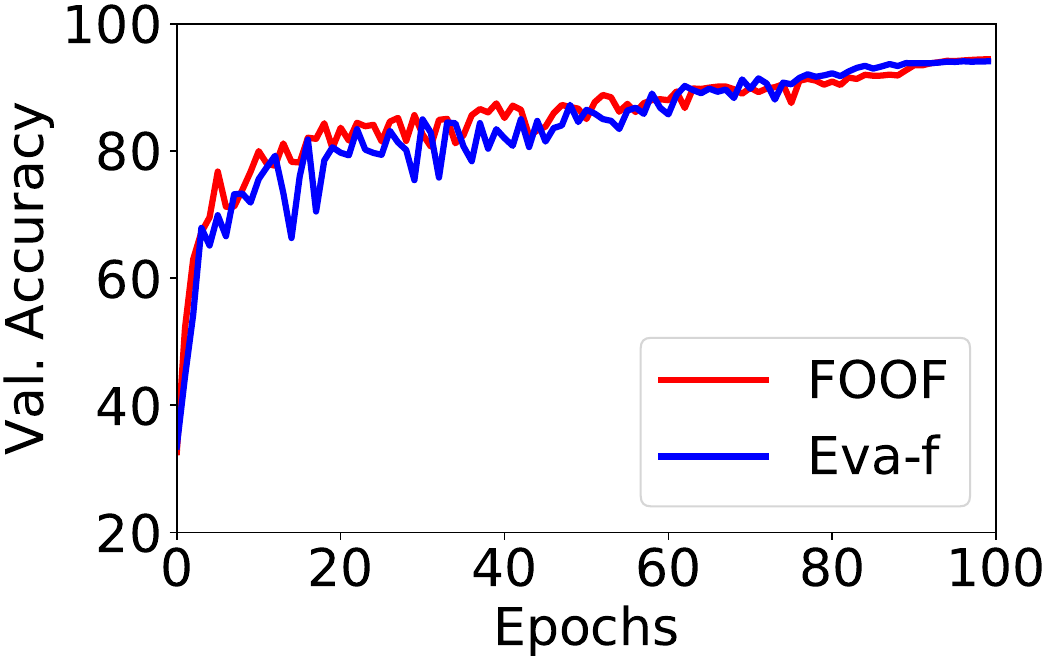}
    }
    \subfloat[Eva-f, Cifar-100]{
        \includegraphics[width=0.45\columnwidth]{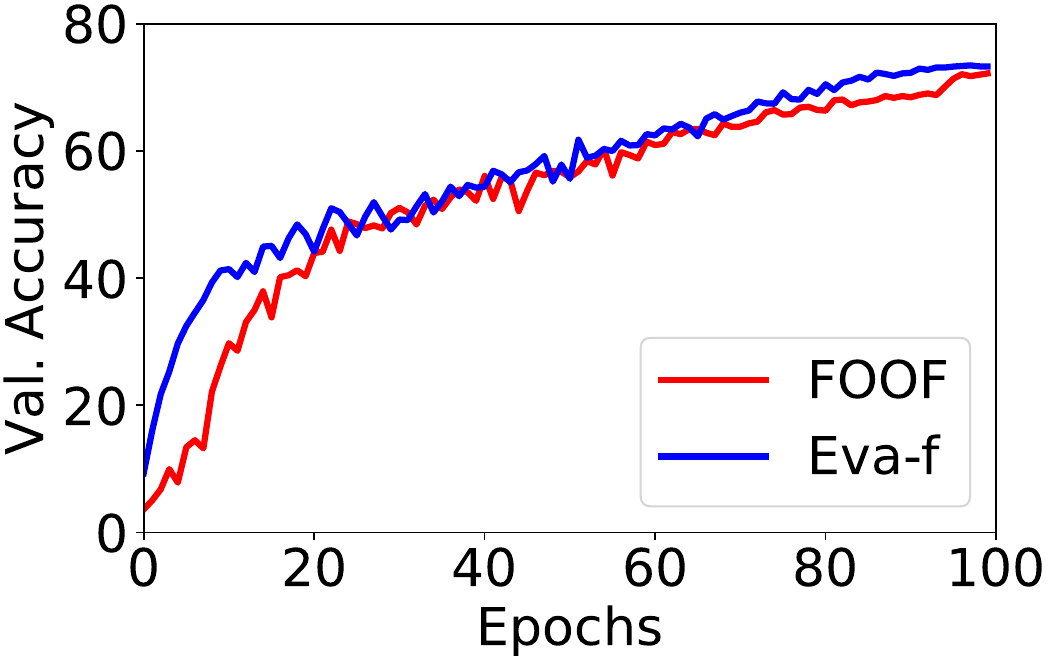}
    } \\
    \subfloat[Eva-s, Cifar-10]{
        \includegraphics[width=0.45\columnwidth]{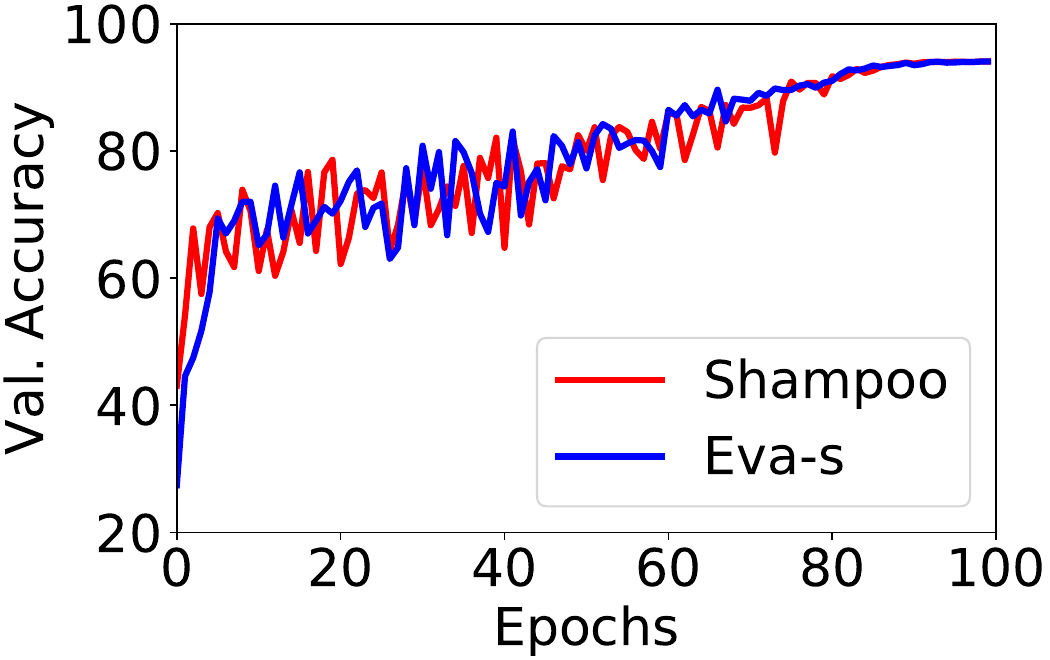}
    }
    \subfloat[Eva-s, Cifar-100]{
        \includegraphics[width=0.45\columnwidth]{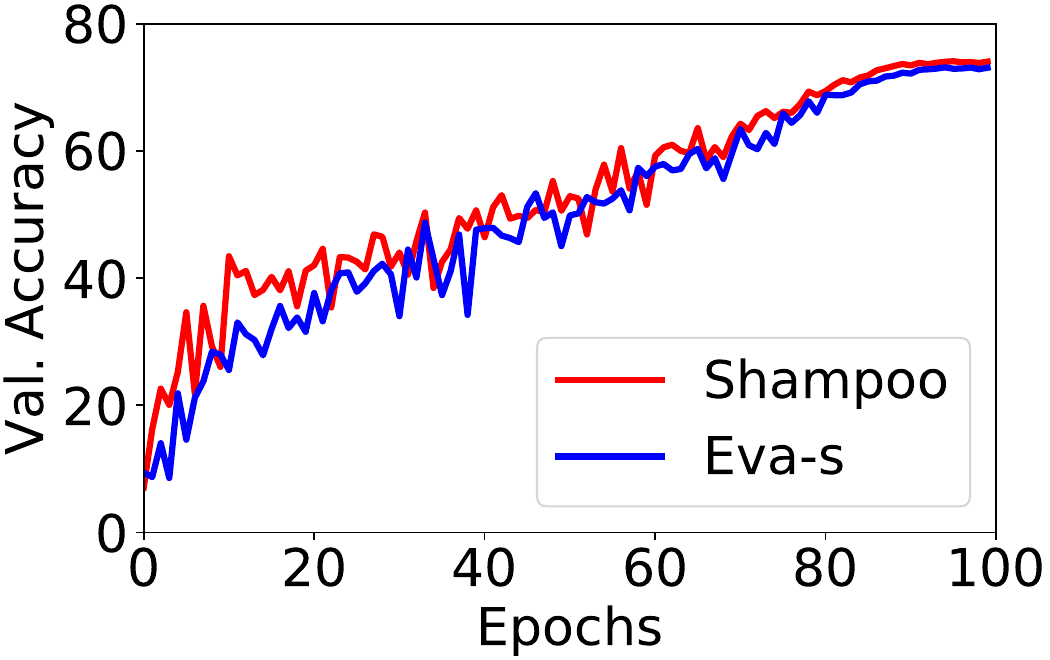}
    }
    \caption{Convergence performance comparison between Eva-f and FOOF, and between Eva-s and Shampoo for training ResNet-110 on Cifar-10 and training VGG-19 on Cifar-100, respectively. }
    \label{fig:convergence-eva-f}
\end{figure}

First of all, we study their convergence performance in training ResNet-110 on Cifar-10 and VGG-19 on Cifar-100 with 100 epochs. We use cosine learning rate schedule and keep other hyper-parameters the same. We report results in Fig.~\ref{fig:convergence-eva-f}. One can see that Eva-f converges similarly to (or even slightly better than) FOOF in the Cifar-10 (or Cifar-100) dataset, and Eva-s converges very closely to Shampoo in both datasets. The similar performance between FOOF and Eva-f, Shampoo and Eva-s, as well as K-FAC and Eva, has validated that our vectorized approximation approach can well maintain the fast convergence performance of original second-order methods on different tasks. 

Second, we report the relative per-iteration time and memory consumption of Eva-f and Eva-s over SGD in Table~\ref{table:efficiency-eva-f}. Though FOOF and Shampoo are compute and memory expensive as we discussed before, their vectorized versions have very low time and memory costs compared to first-order SGD. Specifically, Eva-f and Eva-s only have an average of 1.16$\times$ and 1.35$\times$ of iteration time than SGD, respectively. They are much faster than Shampoo (which are reported in Table~\ref{table:efficiency}), while taking almost the same memory footprint as SGD with vectorized approximation. 

\begin{table}[!ht]
    \centering
     \caption{Relative iteration time and memory of Eva-f and Eva-s over SGD.}
    \label{table:efficiency-eva-f}
    \centering
    \begin{tabular}{cc|cc|cc}
    \hline
  \multirow{2}{*}{Dataset} & \multirow{2}{*}{Model} & \multicolumn{2}{c|}{Eva-f} & \multicolumn{2}{c}{Eva-s} \\ 
  ~ & ~ & Time & Mem & Time & Mem \\ \hline 
  \multirow{3}{*}{Cifar-10} & VGG-19    & 1.10$\times$ & 1.00$\times$ & 1.13$\times$ & 1.00$\times$ \\ 
  & ResNet-110   & 1.11$\times$ & 1.00$\times$ & 1.35$\times$ & 1.00$\times$ \\ 
  & WRN-28-10 & 1.02$\times$ & 1.00$\times$ & 1.04$\times$ & 1.00$\times$ \\ \hline
  \multirow{3}{*}{ImageNet} & ResNet-50    & 1.12$\times$ & 1.00$\times$ & 1.26$\times$ & 1.00$\times$ \\ 
  & Inception-v4 & 1.43$\times$ & 1.00$\times$ & 2.00$\times$ & 1.00$\times$ \\ 
  & ViT-B/16     & 1.20$\times$ & 1.00$\times$ & 1.32$\times$ & 1.00$\times$ \\ \hline
  \end{tabular}
\end{table}

\subsection{Limitation and future work}
Though we have demonstrated the good performance of our proposed Eva, we would like to discuss several limitations and possible future work: (1) there lacks a solid theoretical analysis on the convergence rate of Eva approximations for K-FAC, FOOF, and Shampoo algorithms, (2) since most training tricks were initially proposed for first-order algorithms, it is of interest to design novel second-order friendly strategies for achieving possibly better performance; (3) we will conduct more experiments to show the effectiveness of Eva on other applications such as pre-training large language models; (4) our prototype implementation of Eva currently only supports data parallelism for distributed training, which can be further integrated with model parallelism~\cite{huang2019gpipe,shoeybi2019megatron} for training very large models. 


\section{Related Work}
\textbf{Matrix-free methods} do not explicitly construct second-order matrix, but they rely on the matrix-vector products to calculate the preconditioned gradients. The very initial work in this line is the Hessian-free method~\cite{martens2010deep}, which requires only Hessian-vector products with an iterative conjugate gradient (CG) approach. To reduce the cost of each CG iteration, subsampled mini-batch can be used for Hessian-vector products~\cite{Erdogdu2015subsampled}. Recently, M-FAC~\cite{mfac2021frantar} is proposed to estimate inverse-Hessian vector products with a recursive Woodbury-Sherman-Morrison formula~\cite{amari1998natural}. However, matrix-free methods forgo second-order matrix at a cost of either performing expensive CG iterations or storing extra sliding gradients. 

\textbf{Approximation methods}, on the other hand, construct smaller second-order matrix with different approximation techniques such as quasi-Newton~\cite{Goldfarb20quasi}, quantization~\cite{Alimisis2021CEDO}, Hessian diagonal~\cite{yao2021adahessian,liu2023sophia}, and the most relevant one K-FAC~\cite{martens2015optimizing,grosse2016kronecker}, among which K-FAC is a relatively practical one for deep learning. With K-FAC, one only needs to construct and invert KFs for preconditioning, which is much more efficient than inverting large FIM directly. However, inverting and storing KFs are not cheap enough compared with SGD, thus, many recent works attempt to accelerate K-FAC based on distributed training~\cite{osawa2019large,osawa2020scalable,ueno2020rich,pauloski2020convolutional,pauloski2021kaisa,shi2021accelerating,Zhang2022ScalableKFAC,zhang2023padkfac}. Besides, full-matrix adaptive methods such as Shampoo~\cite{Gupta2018ShampooPS,anil2021scalable} are very similar to K-FAC, which construct second-order preconditioners for gradient tensors on each dimension (concretely, a variant of full-matrix Adagrad). However, the required inverse $p$-th root computations are more expensive than inverting KFs. These expensive memory and compute costs make them even slower than first-order SGD.

\section{Conclusion}
We proposed an efficient second-order algorithm called Eva to accelerate DNN training. We first proposed to use the Kronecker factorization of two small vectors to construct second-order information, which significantly reduces memory consumption. Then we derived a computational-friendly update formula without explicitly calculating the inverse of the second-order matrix using the Sherman–Morrison formula, which reduces the per-iteration computing time. We provided a general vectorized approximation framework to improve two more second-order methods. Extensive experiments were conducted to validate its effectiveness and efficiency, and the results show that Eva outperforms existing popular first-order and second-order algorithms on multiple models and datasets. 

\ifCLASSOPTIONcompsoc
  \section*{Acknowledgments}
\else
  \section*{Acknowledgment}
\fi

The research was supported in part by a RGC RIF grant under the contract R6021-20, a RGC CRF grant under the contracts C7004-22G and C1029-22G, and RGC GRF grants under the contracts 16209120, 16200221 and 16207922.

\ifCLASSOPTIONcaptionsoff
  \newpage
\fi


\bibliographystyle{IEEEtran}
\bibliography{cites}

\begin{IEEEbiography}[{\includegraphics[width=1in,height=1.25in,clip,keepaspectratio]{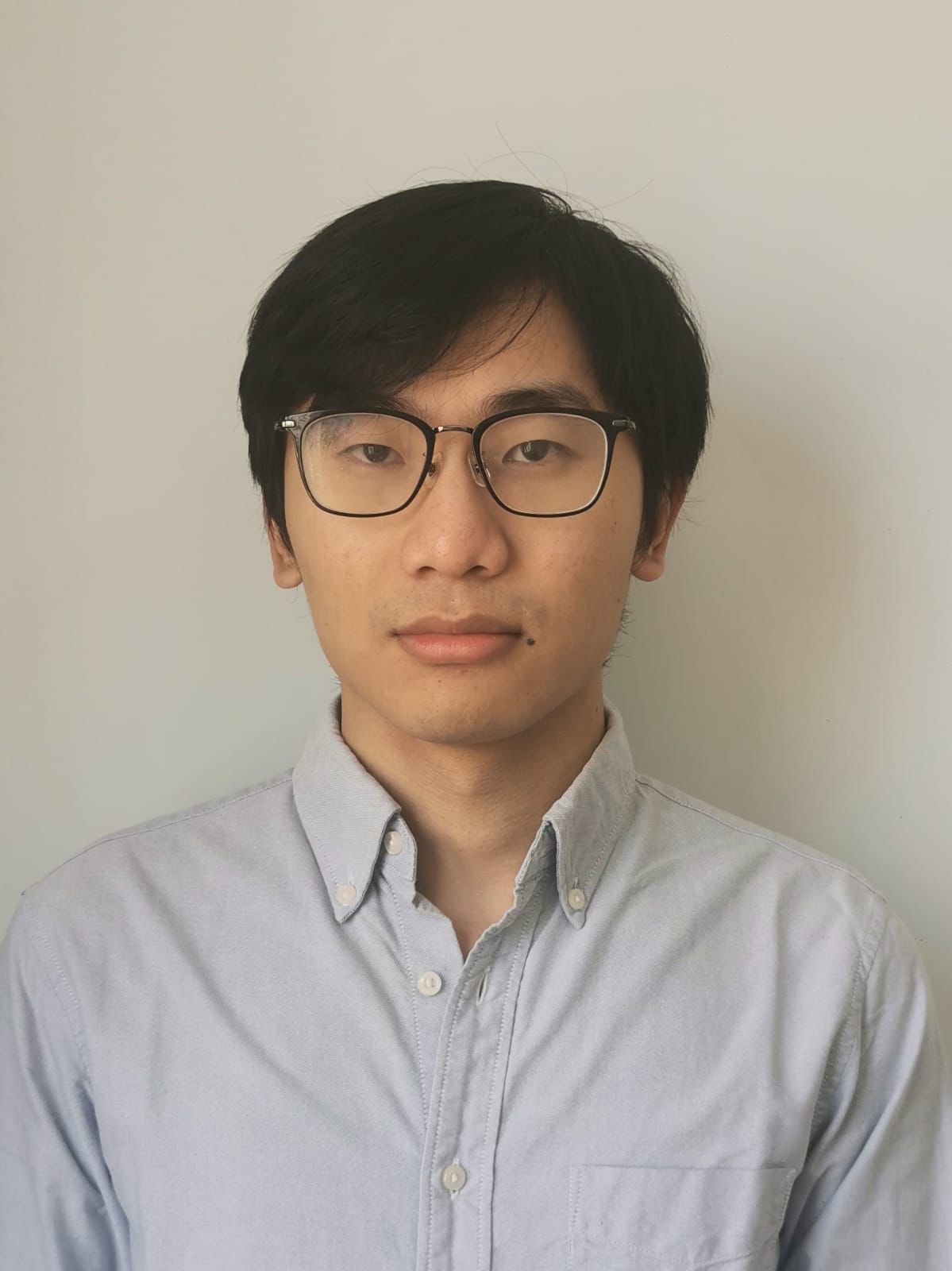}}]{Lin Zhang} received the B.S. degree at School of Electrical Engineering and Automation from Zhejiang University in 2018. He is currently pursuing the Ph.D. degree in the Department of Computer Science and Engineering at the Hong Kong University of Science and Technology. His research interests include machine learning systems and applications, with a special focus on second-order optimization methods, and unsupervised learning on graphs. 
\end{IEEEbiography}

\begin{IEEEbiography}[{\includegraphics[width=1in,height=1.25in,clip,keepaspectratio]{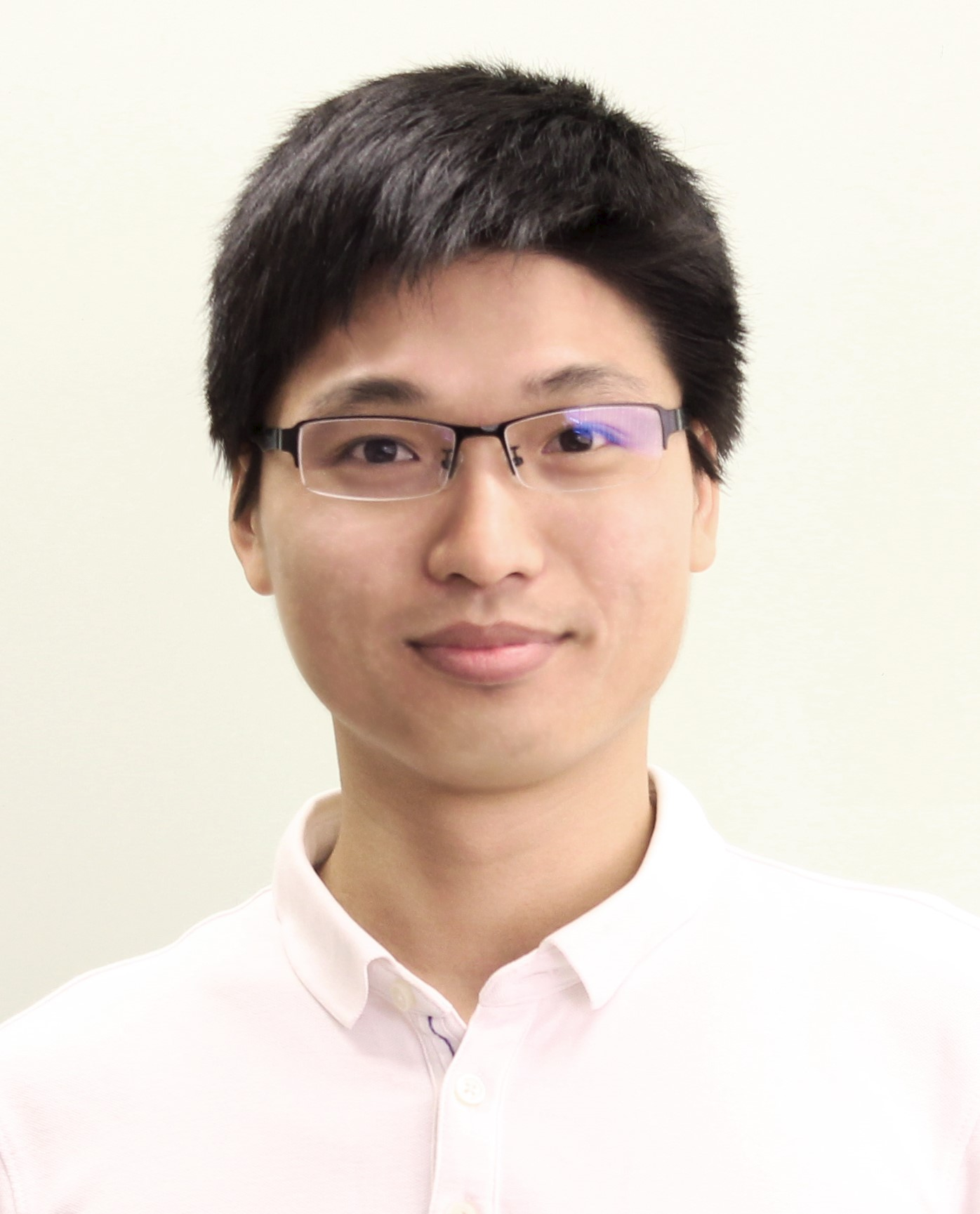}}]{Shaohuai Shi} received a B.E. degree in software engineering from South China University of Technology, P.R. China, in 2010, an MS degree in computer science from Harbin Institute of Technology, P.R. China in 2013, and a Ph.D. degree in computer science from Hong Kong Baptist University in 2020. He is currently an assistant professor in the School of Computer Science and Technology at Harbin Institute of Technology, Shenzhen. His research interests include GPU computing and machine learning systems. He is a member of the IEEE.
\end{IEEEbiography}

\begin{IEEEbiography}[{\includegraphics[width=1in,height=1.25in,clip,keepaspectratio]{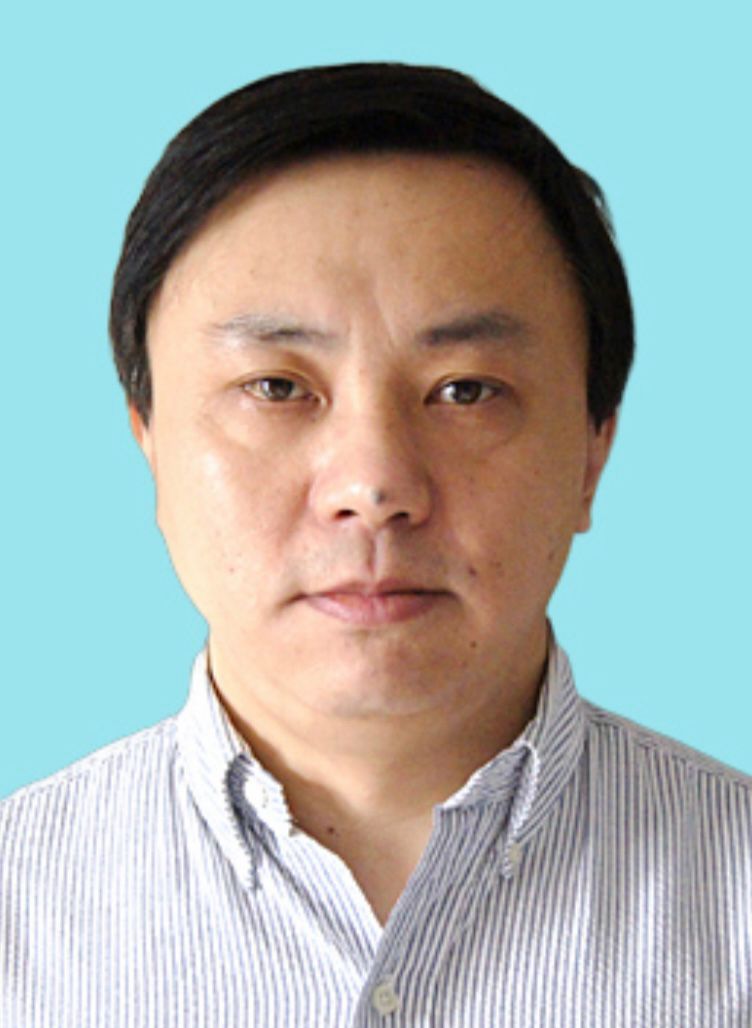}}]{Bo Li} is a professor in the Department of Computer Science and Engineering, Hong Kong University of Science and Technology. He holds the Cheung Kong chair professor in Shanghai Jiao Tong University. Prior to that, he was with IBM Networking System Division, Research Triangle Park, North Carolina. He was an adjunct researcher with Microsoft Research Asia-MSRA and was a visiting scientist in Microsoft Advanced Technology Center (ATC). He has been a technical advisor for China Cache Corp. (NASDAQ CCIH) since 2007. He is an adjunct professor with the Huazhong University of Science and Technology, Wuhan, China. His recent research interests include: large-scale content distribution in the Internet, Peer-to-Peer media streaming, the Internet topology, cloud computing, green computing and communications. He is a fellow of the IEEE for “contribution to content distributions via the Internet”. He received the Young Investigator Award from the National Natural Science Foundation of China (NSFC) in 2004. He served as a Distinguished lecturer of the IEEE Communications Society (2006-2007). He was a corecipient for three Best Paper Awards from IEEE, and the Best System Track Paper in ACM Multimedia (2009).
\end{IEEEbiography}

\end{document}